\RequirePackage{fix-cm}
\documentclass[smallextended]{svjour3}       % onecolumn (second format)
\smartqed  % flush right qed marks, e.g. at end of proof
\usepackage{fullpage}
\usepackage{graphicx}
\usepackage{epstopdf}
\usepackage{graphicx}
\usepackage{hyperref} % Automatically links the label references
\usepackage{amsmath}
\usepackage{algorithm,algorithmic}
\usepackage{colortbl}
\usepackage{color}
\usepackage{amssymb}
\usepackage[numbers]{natbib}
\usepackage{longtable}
\usepackage{subfig}
\usepackage{multirow}
\usepackage{siunitx}

\begin{document}
	
	\title{Stochastic Trust Region Inexact Newton Method for Large-scale Machine Learning}
	%\title{Trust Region Hessian Free Newton with Variance Reduction for Large-Scale Machine Learning%\thanks{Grants or other notes
	%%about the article that should go on the front page should be
	%%placed here. General acknowledgments should be placed at the end of the article.}
	%}
	%\subtitle{Do you have a subtitle?\\ If so, write it here}
	
	\titlerunning{Stochastic Trust Region Inexact Newton Method}        % if too long for running head
	
	\author{Vinod~Kumar~Chauhan
		$\color{red}^*$ \thanks{$\color{red}^*$Most of this work was done when author was doing his PhD at Panjab University Chandigarh, India.\\ $\color{red}^*$This is a pre-print of an article published in International Journal of Machine Learning and Cybernetics. The final authenticated version is available online at: https://doi.org/10.1007/s13042-019-01055-9}         
		\and	
		Anuj~Sharma \and
		Kalpana~Dahiya
	}
	
	%\authorrunning{Short form of author list} % if too long for running head
		
	\institute{Vinod Kumar Chauhan \at
		Department of Engineering\\
		University of Cambridge, UK\\
		%              Tel.: +123-45-678910\\
		%              Fax: +123-45-678910\\
		\email{vk359@cam.ac.uk}\\
		Homepage: \url{http://www.eng.cam.ac.uk/profiles/vk359}          %  \\
		%             \emph{Present address:} of F. Author  %  if needed
		\and
		Anuj Sharma \at
		Computer Science \& Applications\\
		Panjab University Chandigarh, INDIA\\
		\email{anujs@pu.ac.in}\\
		Homepage: \url{https://sites.google.com/view/anujsharma}
		\and
		Kalpana Dahiya \at
		University Institute of Engineering and Technology\\
		Panjab University Chandigarh, INDIA\\
		\email{kalpanas@pu.ac.in}
	}
	
	\date{Received: date / Accepted: date}
	% The correct dates will be entered by the editor

\maketitle

\begin{abstract}
Nowadays stochastic approximation methods are one of the major research direction to deal with the large-scale machine learning problems. From stochastic first order methods, now the focus is shifting to stochastic second order methods due to their faster convergence and availability of computing resources. In this paper, we have proposed a novel Stochastic Trust RegiOn Inexact Newton method, called as STRON, to solve large-scale learning problems which uses conjugate gradient (CG) to inexactly solve trust region subproblem. The method uses progressive subsampling in the calculation of gradient and Hessian values to take the advantage of both, stochastic and  full-batch regimes. We have extended STRON using existing variance reduction techniques to deal with the noisy gradients and using preconditioned conjugate gradient (PCG) as subproblem solver, and empirically proved that they do not work as expected, for the large-scale learning problems. Finally, our empirical results prove efficacy of the proposed method against existing methods with bench marked datasets.
\keywords{stochastic optimisation \and subsampling \and second order methods \and inexact newton \and large-scale learning}
% \PACS{PACS code1 \and PACS code2 \and more}
% \subclass{MSC code1 \and MSC code2 \and more}
\end{abstract}

\section{Introduction}
\label{sec_intro}
\indent Machine learning involves data intensive optimization problems which have large number of component functions corresponding to large amount of available data. Traditional/classical methods, like Newton method, fail to perform well on such large-scale learning problems (i.e., problems with large number of data points) due to large per-iteration complexity. So nowadays one of the major challenge in machine learning is to develop scalable and efficient algorithms to deal with these large-scale learning problems \cite{Zhou2017,Chauhan2017Saag,Chauhan2018Review}.\\
\indent To solve the machine learning problems, gradient descent (GD) \cite{Cauchy1847} is the classical method of choice but it trains slowly while dealing with large-scale learning problems due to high per-iteration cost. Stochastic approximation based methods \cite{Robbins1951} can be quite effective in such situations but they converge slowly due to the noisy approximations of gradient. So a variety of stochastic variance reduction techniques came to existence, e.g.,  \cite{Roux2012,Defazio2014,Schmidt2016,Allen2017,Shang2018}. But the major limitation of these methods is that they can converge up to linear rate only.\\
\indent Newton method is another classical method to solve optimization problems, which can give up to quadratic convergence rate \cite{Boyd2004}. But again (pure) Newton method is not feasible with large-scale learning problems due to huge per-iteration computational complexity and need to store a huge Hessian matrix. So nowadays one of the most significant open question in optimization for machine learning is: ``Can we develop stochastic second order methods with quadratic convergence, like Newton method but has low per-iteration complexity, like stochastic approximation methods?". After success in the stochastic first order methods, the research is shifting its focus towards the stochastic second order methods to leverage the faster convergence of second order methods and the available computing power.\\
\indent Inexact Newton (also called truncated Newton or Hessian free) methods and quasi-Newton methods are among the major research directions for developing second order methods \cite{Bollapragada2016}. Inexact Newton methods try to solve the Newton equation approximately without calculating and storing the Hessian matrix. On the other hand, quasi-Newton methods try to approximate the Hessian inverse and avoid the need to store the Hessian matrix. Thus both the methods try to resolve the issues with Newton method for solving large-scale learning problems. The stochastic variants of inexact Newton and quasi-Newton, further reduce the complexity of these methods by using subsampled gradient and Hessian calculations. In this paper, we have proposed a novel \textbf{s}tochastic \textbf{t}rust \textbf{r}egi\textbf{o}n inexact \textbf{N}ewton (STRON) method to solve the large-scale learning problems, which introduces subsampling to gradient and Hessian calculations. It uses progressive subsampling to enjoy the benefits of both the regimes, stochastic approximation and full-batch learning. We further extend the method using existing variance reduction techniques to deal with noise produced by subsampling of gradient values, and by proposing PCG for solving the trust region subproblem.

\subsection{Optimization Problem}
\label{subsec_opt}
We consider unconstrained convex optimization problem of expected risk, as given below:
\begin{equation}
	\label{eq_expectedrisk}
	\min_{w} R(w) = \mathbb{E}\left[ f\left(w; \xi \right) \right],
\end{equation}
where $f\left(w; \xi \right) = f\left(w; x_i, y_i\right) = f\left(h\left(w; x_i\right), y_i\right)$ is a smooth composition of linear prediction model $h$ and loss function $f$ over randomly selected data point $\left(x_i, y_i\right)$ from the unknown distribution $P\left(x_i, y_i\right)$, parameterized by the model parameter $w \in \mathbb{R}^n$. Since it is not feasible to solve (\ref{eq_expectedrisk}) as $P$ is unknown so the model is approximated by taking a set $N= \left\lbrace (x_1, y_1),...,(x_l,y_l) \right\rbrace$ of $l$ data points from the unknown distribution $P$ and then solving the empirical risk minimization problem, as given below:
\begin{equation}
	\label{eq_erm}
	\min_{w} F(w) = \dfrac{1}{l} \sum_{i=1}^{l} f(w; x_i, y_i).
\end{equation}
For simplicity, we take $f\left(w; x_i, y_i \right) = f_i(w)$. Finite sum optimization problems of type (\ref{eq_erm}) exists across different fields, like signal processing, statistics, operation research, data science and machine learning, e.g., logistic regression and SVM in machine learning.
%In this paper, we consider strongly convex problems because they can be easily obtained by adding l$_2$-regularization.
\subsection{Solution Techniques}
\label{subsec_solutions}
Simple iterative classical method to solve (\ref{eq_erm}) is gradient descent (GD) \cite{Cauchy1847}, as given below:
\begin{equation}
	\label{eq_gd}
	w^{k+1} = w^k - \alpha_k \nabla F\left(w_k\right),
\end{equation}
where $(k+1)$ is iteration number and $\alpha_k$ is called learning rate or step size. The complexity of this iteration is $O\left(nl\right)$ which is large for large-scale learning problems due to large number of data points. Stochastic gradient descent (SGD) \cite{Robbins1951} is very effective to deal with such problems due to its low per-iteration complexity, as given below:
\begin{equation}
	\label{eq_sgd}
	w^{k+1} = w^k - \alpha_k \nabla f_{i_k}\left(w_k\right),
\end{equation}
where $i_k$ is randomly selected data point. But convergence can't be guaranteed in SGD because of noisy gradient values.\\
\indent Another classical second order method to solve (\ref{eq_erm}) is Newton method, as given below:
\begin{equation}
	\label{eq_newton}
	w^{k+1} = w^k - \alpha_k \nabla^2 F\left(w_k\right)^{-1} \nabla F\left(w_k\right).
\end{equation}
The complexity of this iteration is $O(n^2l + n^3)$ and it needs to store and invert the Hessian matrix $\nabla^2 F\left(w_k\right)$, which is computationally very expensive and needs large memory for large-scale learning problems, respectively. That's why first order methods and their stochastic variants have been studied very extensively, during the last decade, to solve the large-scale learning problems but not second order methods. As stochastic first order methods have hit their limits and due to good availability of computing power, the main focus is shifting towards stochastic second order methods, and nowadays one important open question is to find stochastic second order methods with quadratic convergence rates.\\
\indent There are two major research directions for second order methods: quasi-Newton methods and inexact Newton methods, both of which try to resolve the issues associated with the Newton method. Quasi-Newton methods try to approximate the Hessian matrix during each iteration, as given below:
\begin{equation}
	\label{eq_quasi-newton}
	w^{k+1} = w^k - \alpha_k B_k \nabla F\left(w_k\right),
\end{equation}
where $B_k$ is an approximate of $\nabla^2 F\left(w_k\right)^{-1}$, e.g., Broyden-Fletcher-Goldfarb-Shanno (BFGS) algorithm is one such method \cite{Fletcher1980}. On the other hand, inexact Newton methods try to solve the Newton system approximately, e.g., Newton-CG \cite{Steihaug1983}. Both the methods try to resolve the issues related with Newton method but still their complexities are large for large-scale problems. So a number of stochastic variants of these methods have been proposed, e.g., \cite{Byrd2011,Byrd2016,Bollapragada2018,Bellavia2018} which introduce subsampling to gradient and Hessian calculations.
%In this paper, we have proposed a novel \textbf{s}tochastic \textbf{t}rust \textbf{r}egi\textbf{o}n inexact \textbf{N}ewton (STRON) method which introduces subsampling to gradient and Hessian calculations. It uses progressive subsampling to enjoy the benefits of both the regimes, stochastic approximation and full batch processing. We further extend the method using existing variance reduction techniques to deal with noise produced by subsampling of gradient values, and by proposing PCG for solving the trust region subproblem.

\subsection{Contributions}
\label{subsec_contributions}
The contributions of the manuscript are listed below:

\begin{itemize}
	\item The manuscript highlights the recent shift of stochastic methods from first order to second order methods and raises an open question that we need stochastic second order methods with quadratic convergence rate to solve the large-scale machine learning problems.
	
	\item We have proposed a novel subsampled variant of trust region inexact Newton method for solving large-scale learning problems, which is called STRON and is the first stochastic variant of trust region inexact Newton methods. STRON uses progressive subsampling scheme for gradient and Hessian calculations to enjoy the benefits of both stochastic and full batch regimes. STRON can converge up to quadratic rate and answers the raised open question.
	
	\item STRON has been extended using existing variance reduction techniques to deal with the noisy approximations of the gradient calculations. The extended method uses stochastic variance reduced gradient (SVRG) for variance reduction with static batching for gradient calculations and progressive batching for the Hessian calculations. The empirical results prove that this does not work as expected for large-scale learning.
	
	\item We further extend STRON and use PCG, instead of CG method, to solve the Newton system inexactly. We have used weighted average of identity matrix and diagonal matrix as the preconditioner. But even this fails to work as expected for large-scale learning.
	
	\item Finally, our empirical experiments prove the efficacy of STRON against existing techniques with bench marked datasets.
\end{itemize}

\section{Literature Review}
\label{sec_literature}
Stochastic approximation methods are very effective to deal with the large-scale learning problems due to their low per-iteration cost, e.g., SGD \cite{Robbins1951}, but they lead to slow convergence rates due to the noisy approximation. To deal with the noise issue, a number of techniques have been proposed and some of the important techniques (as discussed in \cite{Csiba2016}) are: (a) decreasing learning rates \cite{Shalev-Shwartz2007}, (b) using mini-batching \cite{Chauhan2018SS_AI}, (c) importance sampling \cite{Csiba2016}, and (d) variance reduction \cite{Johnson2013}. The variance reduction methods can further be classified into three categories: primal methods \cite{Schmidt2016,Chauhan2019SAAGs34}, dual methods \cite{Shalev2013a} and primal-dual methods \cite{Zhang2015}. The variance reduction techniques are effective to deal with the large-scale learning problems because of low per-iteration complexity, like SGD, and have fast linear convergence, like GD. These techniques exploit the best of SGD and GD but for these stochastic variants of first order methods the convergence is limited to linear rate only, unlike the second order methods which can give up to quadratic rate.\\
%This issue is fixed using following major techniques: (i) using mini-batching \cite{Yang2018}, (ii) importance sampling \cite{Csiba2016}, (iii) variance reduction techniques \cite{Johnson2013}, and (iv) decreasing step sizes \cite{Shalev-Shwartz2007}, (see, \cite{Csiba2016} for details). Variance reduction techniques can be classified into three categories: primal methods \cite{Johnson2013,Schmidt2016}, dual methods \cite{Shalev2013a} and primal-dual methods \cite{Zhang2015}. SVRG \cite{Johnson2013} is one of the most widely used variance reduction technique, which has been extended even to parallel \& distributed settings and also used in second order methods \cite{Zhang2018}. These techniques are effective to deal with the large-scale learning problems because of low per-iteration complexity, like SGD and have fast linear convergence like GD. But the variance reduction techniques are limited to linear convergence, far away from quadratic convergence of second order methods.\\
\indent Second order methods utilize the curvature information to guide the step direction towards the solution and exhibit faster convergence than the first order methods. But huge per-iteration cost due to the huge Hessian matrix and its inversion make the training of models slow for large-scale problems. So certain techniques have been developed to deal with the issues related to Hessian matrix, e.g., quasi-Newton methods and inexact Newton methods are two major directions to deal with the huge computational cost of Newton method. Quasi-Newton methods approximate the Hessian matrix during each iteration, e.g., BFGS \cite{Fletcher1980} and its limited memory variant, called L-BFGS \cite{Liu1989}, are examples of the quasi-Newton class which use gradient and parameter values from the previous iterations to approximate the Hessian inverse. L-BFGS uses only recent information from previous $M$-iterations. On the other hand, inexact Newton methods try to solve the Newton system approximately, e.g., Newton-CG \cite{Steihaug1983}.\\
\indent Recently, several stochastic variants of BFGS and L-BFGS have been proposed to deal with large-scale problems.
%e.g., \cite{Schraudolph2007,Byrd2011,Byrd2016,Bollapragada2016}. 
\citet{Schraudolph2007} proposed stochastic variants of BFGS and L-BFGS for the online setting, known as oBFGS. \citet{Mokhtari2014} extended oBFGS by adding regularization which enforces upper bound on the eigen values of the approximate Hessian, known as RES (Regularized Stochastic BFGS). Stochastic quasi-Newton (SQN) \cite{Byrd2016} is another stochastic variant of L-BFGS which collects curvature information at regular intervals, instead of at each iteration. Variance-reduced Stochastic Newton (VITE) \cite{Lucchi2015} extended RES and proposed to use variance reduction for the subsampled gradient values for solving smoothly strongly convex problems. \citet{Kolte2015} provided another stochastic L-BFGS method with variance reduction using SVRG (referred as SVRG-LBFGS). \citet{Moritz2016} proposed Stochastic L-BFGS (SLBFGS) using SVRG for variance reduction and using Hessian-vector product to approximate the gradient differences for calculating the Hessian approximations, also referred as SVRG-SQN. SVRG-LBFGS and SVRG-SQN differ in how the curvature is approximated (i.e. how Hessian is approximated), former collects the curvature information once during each epoch (outer iterations) using gradient differences but later collects the curvature information after regular intervals inside the inner-iterations using Hessian-vector products. \citet{Berahas2016} proposed multi-batch scheme into stochastic L-BFGS where batch sample changes with some overlaps with previous iteration. \citet{Bollapragada2018} proposed progressive batching, stochastic line search and stable Newton updates for L-BFGS. \citet{Bollapragada2016} studies the conditions on the subsample sizes to get the different convergence rates.\\
%Some other examples of stochastic quasi-Newton methods are \cite{}.\\
\indent Stochastic inexact Newton methods are also explored extensively. \citet{Byrd2011} proposed stochastic variants of Newton-CG along with L-BFGS method. \citet{Bollapragada2016} studies subsampled Newton methods and find conditions on subsample sizes and forcing term (constant used with the residual condition), for linear convergence of Newton-CG method. \citet{Bellavia2018} studies the effect of forcing term and line search to find linear and super-linear convergence of Newton-CG method. Newton-SGI (stochastic gradient iteration) is another way of solving the linear system approximately and is studied in \citet{Agarwal2017}.\\
\indent Trust Region Newton (TRON) method is one of the most efficient solver for solving large-scale linear classification problems \cite{TRON}. This is trust region inexact Newton method which does not use any subsampling and is present in LIBLINEAR library \cite{LIBLINEAR}. \citet{Hsia2017} extends TRON by improving the trust region radius value. \citet{Hsia2018}, further extends TRON and uses preconditioned conjugate gradient (PCG) which uses weighted average of identity matrix and diagonal matrix as a preconditioner, to solve the trust region subproblem. Since subsampling is an effective way to deal with the large-scale problems so in this paper, we have proposed a stochastic variant of trust region inexact Newton method, which have not been studied so far to the best of our knowledge.

\section{Trust Region Inexact Newton Method}
\label{sec_tron}
Inexact Newton methods, also called as Truncated Newton or Hessian free methods, solve the Newton equation (linear system) approximately. CG method is a commonly used technique to solve the trust region subproblem approximately. In this section, we discuss inexact Newton method and its trust region variation.

\subsection{Inexact Newton Method}
\label{subsec_IN}
The quadratic model $m_k(p)$ obtained using Taylor's theorem is given below:
\begin{equation}
	\label{eq_model}
	\begin{split}
		F(w_k + p) - F(w_k) \approx m_k(p) \equiv \nabla F\left(w_k\right)^T p + \dfrac{1}{2} p^T \nabla ^2 F\left(w_k\right) p.
	\end{split}
\end{equation}
Taking derivative of $m_k(p)$ w.r.t. $p$ and equating to zero, we get,
\begin{equation}
	\label{eq_newton_equation}
	\begin{split}
		\nabla ^2 F\left(w_k \right) p = - \nabla F\left(w_k\right),
	\end{split}
\end{equation}
which is Newton system and its solution gives Newton method, as given below:
\begin{equation}
	\label{eq_newton2}
	\begin{split}
		w^{k+1} = w^k + p_k = w^k - \nabla^2 F\left(w_k\right)^{-1} \nabla F\left(w_k\right).
	\end{split}
\end{equation}
The computational complexity of this iteration is $O(n^2l + n^3)$ which is very expensive. This iteration involves the calculation and inversion of a large Hessian matrix which is not only very expensive to calculate but expensive to store also. CG method approximately solves the subproblem (\ref{eq_newton_equation}) without forming the Hessian matrix, which solves the issues related to large computational complexity and need to store the large Hessian matrix. Each iteration runs for a given number of CG iterations or until the residual condition is satisfied, as given below:
\begin{equation}
	\label{eq_residual_cg}
	\| r_k \| \le \eta_k^{'} \|\nabla F\left(w_k\right)\|,
\end{equation}
where $ r_k = \nabla ^2 F\left(w_k\right) p + \nabla F\left(w_k\right)$ and $\eta_k^{'}$ is a small positive value, known as forcing term \cite{Nocedal1999}.

\subsection{Trust Region Inexact Newton Method}
\label{subsec_TCIN}
Trust region is a region in which the approximate quadratic model of the given function gives correct approximation for that function. In trust region methods, we don't need to calculate the step size (also called learning rate) directly but they indirectly adjust the step size as per the trust region radius. Trust region method solves the following subproblem to get the step direction $p_k$:
\begin{equation}
	\label{eq_tr_prob}
	\min_{p} \; m_k\left(p\right) \quad \text{s.t.} \quad \|p\| \le \triangle_k,
\end{equation}
where $m_k(p)$ is a quadratic model of $F\left(w_k + p\right) - F\left(w_k\right)$, as given in (\ref{eq_model}) and $\triangle_k$ is the trust region radius. This subproblem can be solved similar to Newton-CG, except that now we need to take care of the extra constraint of $p$. TRON (trust region Newton method) \cite{TRON} is one of the most famous and widely used such method, which is used in LIBLINEAR \cite{LIBLINEAR} to solve l$_2$-regularized logistic regression and l$_2$-SVM problems. \citet{Hsia2017} extends TRON by proposing better trust region radius. \citet{Hsia2018} further extends TRON using PCG subproblem solver which uses average of identity matrix and  diagonal matrix as preconditioner, to solve the trust region subproblem and shows that PCG could be effective to solve ill-conditioned problems.\\
\indent Then the ratio of actual and predicted reductions of the model is calculated, as given below:
\begin{equation}
	\label{eq_ratio}
	\rho_k = \dfrac{F(w_k + p_k) - F(w_k)}{m_k(p_k)}.
\end{equation}
The parameters are updated for the $(k+1)$th iteration as given below:
\begin{equation}
	\label{eq_parameter_update}
	w_{k+1} = \begin{cases}
		w_k + p_k, & \text{if} \; \rho_k > \eta_0,\\
		w_k, & \text{if} \; \rho_k \le \eta_0,
	\end{cases}
\end{equation}
where $\eta_0 > 0$ is a given constant. Then the trust region radius $\triangle_k$ is updated as per the ratio of actual reduction and predicted reduction, and a framework for updating $\triangle_k$ as given in \cite{Lin1999}, is given below:
\begin{equation}
	\label{eq_trust_region}
	\triangle_{k+1} \in
	\begin{cases}
		\left[ \gamma_1 \min\lbrace\|p_k\|, \triangle_k \rbrace, \gamma_2\triangle_k \right], & \text{if}\; \rho_k \le \eta_1,\\
		\left[ \gamma_1 \triangle_k , \gamma_3 \triangle_k \right], & \text{if}\; \rho_k \in \left(\eta_1, \eta_2 \right),\\
		\left[ \triangle_k , \gamma_3 \triangle_k \right], & \text{if}\; \rho_k \ge \eta_2,		
	\end{cases}
\end{equation}
where $0 < \eta_1 < \eta_2 \le 1$ and $ 0 < \gamma_1 < \gamma_2 < 1 < \gamma_3$. If $\rho_k \le \eta_1$ then the Newton step is considered unsuccessful and the trust region radius is shrunk. On the other hand if $\rho_k \ge \eta_2$ then the step is successful and the trust region radius is enlarged. We have implemented this framework as given in the LIBLINEAR library \cite{LIBLINEAR} and chose the following pre-defined values for the above constants: $\eta_0 = 1e-4$, $\eta_1 = 0.25$, $\eta_2 = 0.75$, $\gamma_1 = 0.25$, $\gamma_2 = 0.5$ and $\gamma_3 = 4$.

\section{STRON}
\label{sec_stron}
STRON exploits the best of both, stochastic and batch regimes, using progressive subsampling to solve the large-scale learning problems. As stochastic gradient descent (SGD) trains faster for large-scale learning problems than gradient descent (GD) due to low computations per iteration but GD is more accurate than SGD due to batch calculations, similarly STRON takes benefit of low computation during initial iterations and as it reaches the solution region it uses batch calculations to find accurate solution like TRON. The major challenge with STRON is to decide when to switch from stochastic to full-batch regime, i.e., to tune the subsampling rate.\\
\indent STRON introduces stochasticity into the trust region inexact Newton method and calculates subsampled function, gradient and Hessian values to solve the trust region subproblem, as given below:
\begin{equation}
	\label{eq_stron}
	\begin{split}
		\min_{p}\; m_k(p) = \nabla F_{X_k}\left(w_k\right)^T p + \dfrac{1}{2} p^T \nabla ^2 F_{S_k}\left(w_k\right) p, \quad \text{s.t.} \quad \|p\| \le \triangle_k,
	\end{split}
\end{equation}
where $\nabla ^2 F_{S_k}\left(w_k \right)$ and $\nabla F_{X_k}\left(w_k\right) $ are subsampled Hessian and gradient values over the subsamples $S_k$ and $X_k$, respectively, as defined below:
\begin{equation}
	\label{eq_sub_grad_Hess}
	\begin{split}
		\nabla ^2 F_{S_k}\left(w_k \right) = \dfrac{1}{|S_k|} \sum_{i \in S_k} \nabla ^2 f_i\left(w_k \right),\\
		\nabla F_{X_k}\left(w_k \right) = \dfrac{1}{|X_k|} \sum_{i \in X_k} \nabla f_i\left(w_k \right),\\
		F_{X_k}\left(w_k \right) = \dfrac{1}{|X_k|} \sum_{i \in X_k} f_i\left(w_k \right),
	\end{split}
\end{equation}
where subsamples are increasing, i.e., $ |X_{k}| < |X_{k+1}|$, $ |S_{k}| < |S_{k+1}|$ and $F_{X_k}$ is subsampled function value used for calculating $\rho_k$. STRON solves (\ref{eq_stron}) approximately for given number of CG iterations or until the following residual condition is satisfied:
\begin{equation}
	\label{eq_strong_residual_cg}
	\|r_k\| \le \eta_k^{'} \|\nabla F_{X_k}\left(w_k\right)\|,
\end{equation}
where $r_k = \nabla ^2 F_{S_k}\left(w_k\right) p + \nabla F_{X_k}\left(w_k\right) $.\\
\begin{algorithm}[htb]
	\caption{STRON}
	\label{algo_stron}	
	\begin{algorithmic}[1]
		\STATE \textbf{Input:} $w_0$\\
		\STATE \textbf{Result:} $w = w_k$
		\FOR{$k=0,1,...$}
		\STATE Randomly select subsamples $S_k$ and $X_k$
		\STATE Calculate subsampled gradient $\nabla F_{X_k}(w_k)$
		\STATE Solve the trust region subproblem using Algorithm~\ref{algo_cg}, to get the step direction $p_k$
		\STATE Calculate the ratio $\rho_k = \left(F_{X_k}(w_k + p_k) - F_{X_k}(w_k)\right)/m_k(p_k)$
		\STATE Update the parameters using (\ref{eq_parameter_update})
		\STATE Update the trust region radius $\triangle_k$ using (\ref{eq_trust_region})
		\ENDFOR
	\end{algorithmic}
\end{algorithm}
\begin{algorithm}[htb]
	\caption{CG Subproblem Solver}
	\label{algo_cg}
	\begin{algorithmic}[1]
		\STATE \textbf{Inputs:} $\triangle_k > 0$, $\eta_k^{'} \in(0, 1)$\\
		\STATE \textbf{Result:} $p_k = p_j$
		\STATE Initialize $p_0 = 0, r_0 = d_0 = - \nabla F_{X_k}(w_k)$
		%		\STATE $rTr = r_0^Tr_0$
		\FOR{$j=1,2,...$}
		\IF{$\|r_{j-1}\| < \eta_k^{'} \|\nabla F_{X_k}(w_k)\|$}
		\STATE return $p_k = p_{j-1}$
		\ENDIF
		\STATE Calculate subsampled Hessian-vector product $v_j = \nabla^2 F_{S_k} (w_k) d_{j-1}$
		\STATE $\alpha_j = \|r_{j-1}\|^2/\left(d_{j-1}^T v_j \right)$
		\STATE $p_j = p_{j-1} +\alpha_j d_{j-1}$
		\IF{$\|p_j\| \ge \triangle_k$}
		%		\STATE $p_j = p_{j-1} - \alpha_j d_{j-1}$
		\STATE Calculate $\tau_j$ such that $\|p_{j-1} + \tau_j d_{j-1}\| = \triangle_k$
		\STATE return $p_k = p_{j-1} + \tau_j d_{j-1}$
		\ENDIF
		\STATE $r_j = r_{j-1} - \alpha_j v_j$, %$rTr\_new = r_j^T r_j$
		\STATE $\beta_j = \|r_{j}\|^2/\|r_{j-1}\|^2, d_j = r_j + \beta_j d_{j-1}$
		%		\STATE $rTr = rTr\_new$
		\ENDFOR
	\end{algorithmic}
\end{algorithm}
STRON is presented by Algorithm~\ref{algo_stron}. It randomly selects subsamples $S_k$ and $X_k$ for the $k$th iteration (outer iterations). $X_k$ and $S_k$ are used for calculating the gradient and Hessian values, respectively. Then it solves the trust region subproblem using CG solver (inner iterations) which uses subsampled Hessian in calculating Hessian-vector products. CG stops when residual condition, same as (\ref{eq_strong_residual_cg}), satisfies, it reaches maximum \#CG iterations or it reaches the trust region boundary. The ratio of reduction in actual and predicted reduction is calculated similar to (\ref{eq_ratio}) but using subsampled function, and is used for updating the parameters as given in (\ref{eq_parameter_update}). Then trust region radius $\triangle_k$ is updated as per $\rho_k$ as given in (\ref{eq_trust_region}) and these steps are repeated for a given number of iterations or until convergence.\\
\indent STRON uses progressive subsampling, i.e., dynamic subsampling to calculate function, gradient and Hessian values, and solves the Newton system approximately. It is effective to deal with large-scale problems since it uses subsampling and solves the subproblem approximately, without forming the Hessian matrix but using only Hessian-vector products. So it handles the complexity issues related with the Newton method.

\subsection{Complexity}
\label{subsec_complexity}
The complexity of trust region inexact Newton (TRON) method depends on function, gradient and CG subproblem solver. This is dominated by CG subproblem solver and is given by
$
O\left(nl\right) \times \#\text{CG iterations},
\text{ and for sparse data, } O\left(\#nnz \right) \times \#\text{CG iterations},
$
where $\#nnz$ is number of non-zeros values in the dataset. For subsampled trust region inexact Newton (STRON) method, the complexity per-iteration is given by
$O\left(n|S_k|\right) \times \#\text{CG iterations},
\text{ and for sparse data, } O\left(\#nnz_{S_k} \right) \times \#\text{CG iterations},
$
where $\#nnz_{S_k}$ is number of non-zeros values in the subsample $S_k$. Since \#CG iterations taken by TRON and STRON do not differ much so the per-iteration complexity of STRON is smaller than TRON in the initial iterations and later becomes equal to TRON due to progressive subsampling, i.e., when $|S_k| = N$.

\subsection{Analysis}
\label{sec_analysis}
STRON method uses progressive subsampling with the assumption that eventually mini-batch size becomes equal to the whole dataset, i.e., for some value of $k\ge \bar{k}>0$, STRON becomes TRON. So we can follow the theoretical results given in TRON, which itself refers the results from \cite{Lin1999}. For $k\ge \bar{k}$, we get different convergence depending upon the value of $\eta_k^{'}$: If $\eta_k^{'}<1$ then STRON converges Q-linearly, if $\eta_k^{'} \rightarrow 0$ as $k\rightarrow \infty$ then STRON has Q-superlinear convergence and when $\eta_k^{'} \le \kappa_0 \|\nabla F(w_k)\|$ for $\kappa_0 > 0$ then STRON converges at quadratic rate.

\section{Experimental Results}
\label{sec_exps}
In this section, we discuss experimental settings and results. The experiments have been conducted with the bench marked binary datasets as given in the Table~\ref{tab_datasets}, which are available for download from LibSVM website\footnote{\url{https://www.csie.ntu.edu.tw/\~cjlin/libsvmtools/datasets/}}.
\begin{table}[htb]
	\centering
	\caption{Datasets used in experimentation}
	\label{tab_datasets}
	\begin{tabular}{|l|r|r|r|}
		\hline
		Dataset & \#features & \#datapoints\\
		\hline
		%		HIGGS & 2 &  28 & 11,000,000\\
		%		SUSY & 2 & 18 & 5,000,000\\
		%		SensIT Vehicle & 3 & 100 & 78,823\\
		%		(combined) & & &\\
		%		mnist & 10 & 780 & 60,000\\
		%		protein & 3 & 357 & 17,766\\
		gisette &  5000 & 6,000\\
		rcv1.binary & 47,236  & 20,242\\
		webspam (unigram) & 254 & 350,000\\
		covtype.binary & 54 & 581,012\\
		ijcnn1  & 22 & 49,990\\
		news20.binary & 1,355,191 & 19,996\\
		real-sim  & 20,958  & 72,309\\
		Adult  & 123  & 32,561\\
		mushroom  & 112  & 8124\\
		%		SVHN & 10 & 3,072 & 73,257\\
		\hline
	\end{tabular}
\end{table}
We use following methods in experimentation:\\
\textbf{TRON} \cite{Hsia2018}: This is a trust region inexact Newton method without any subsampling. It uses preconditioned CG method to solve the trust region subproblem and it is present in the current version of LIBLINEAR library \cite{LIBLINEAR}.\\
\textbf{STRON}: This is the proposed stochastic trust region inexact Newton method with progressive subsampling technique for gradient and Hessian calculations. It uses CG method to solve the trust region subproblem.\\
\textbf{STRON-PCG}: This is an extension of STRON using PCG for solving the trust region subproblem, as discussed in the Subsection~\ref{subsec_pcg}.\\
\textbf{STRON-SVRG}: This is another extension of STRON using variance reduction for subsampled gradient calculations, as discussed in Subsection~\ref{subsec_svrg}.\\
\textbf{Newton-CG} \cite{Byrd2011}: This is stochastic inexact Newton method which uses CG method to solve the subproblem. It uses progressive subsampling similar to STRON.\\
\textbf{SVRG-SQN} \cite{Moritz2016}: This is stochastic L-BFGS method with variance reduction for gradient calculations.\\
\textbf{SVRG-LBFGS} \cite{Kolte2015}: This is another stochastic L-BFGS method with variance reduction. It differs from SVRG-SQN method in approach by which Hessian information is sampled.

\subsection{Experimental Setup}
\label{subsec_exp_setup}
The datasets have been divided into 80\% and 20\%, for training and testing datasets, respectively to plot the convergence, and 5-fold cross-validation has been used to present results in Table~\ref{tab_time}. We have used $\lambda = 1/l$ for all methods because generally it gives good results and at the same time help to reduce number of tunable hyper-parameters. \#CG iterations has been set to a sufficiently large value of 25, as all the inexact Newton methods use 5-10 iterations and hardly go beyond 20 iterations. Progressive batching scheme uses initial batch size of 1\% for all datasets except ijcnn which uses 10\% of dataset and the subsample size is increased linearly for all datasets, except covtype where exponential rate is used. Moreover we set the rate such that subsample size equals dataset size in 5 epochs because, generally second order methods reach the solution region in 4-5 epochs and converge in 8-10 epochs. For the sake of simplicity and avoid extra sampling, we take same subsample for Hessian and gradient calculations, i.e, $S_k=X_k$. Quasi-Newton methods (SVRG-SQN and SVRG-LBFGS) use mini-batch size of 10\% with stochastic backtracking line search to find the step size and same size mini-batches are taken for gradient and Hessian subsampling. Memory of $M=5$ is used in quasi-Newton methods with $L=5$ as update frequency of Hessian inverse approximation for SVRG-SQN method. All the algorithms use similar exit criterion of decrease in gradient value where we calculate gradient ($\|g_0\|$) at $w_0$ and run the algorithms until $\|g_k\| \le \epsilon \|g_0\|$, where $\|g_k\|$ is gradient value at k$^{th}$-iteration and $\epsilon$ is the given tolerance level. Moreover all the methods are implemented in C++\footnote{experimental results can be reproduced using the library \href{https://github.com/jmdvinodjmd/LIBS2ML}{LIBS2ML}~\cite{LIBS2ML}.} with MATLAB interface and experiments have been executed on MacBook Air (8 GB 1600 MHz DDR3, 1.6 GHz Intel Core i5 and 256GB SSD).

\subsection{Comparative Study}
\label{subsec_comparisons}
\begin{figure}[htb]
	\subfloat{\includegraphics[width=.5\linewidth,height=3.6cm]{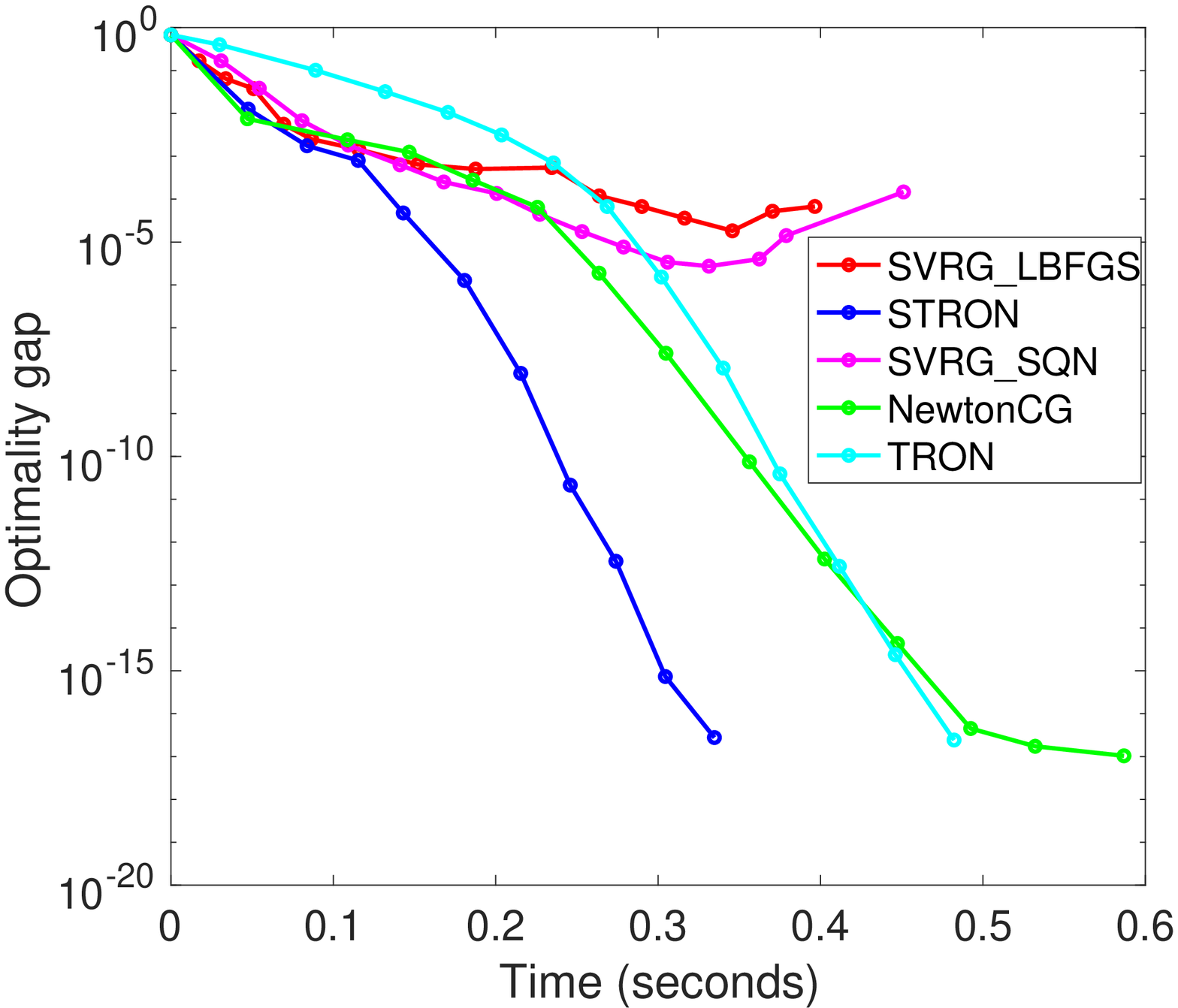}}
	\subfloat{\includegraphics[width=.5\linewidth]{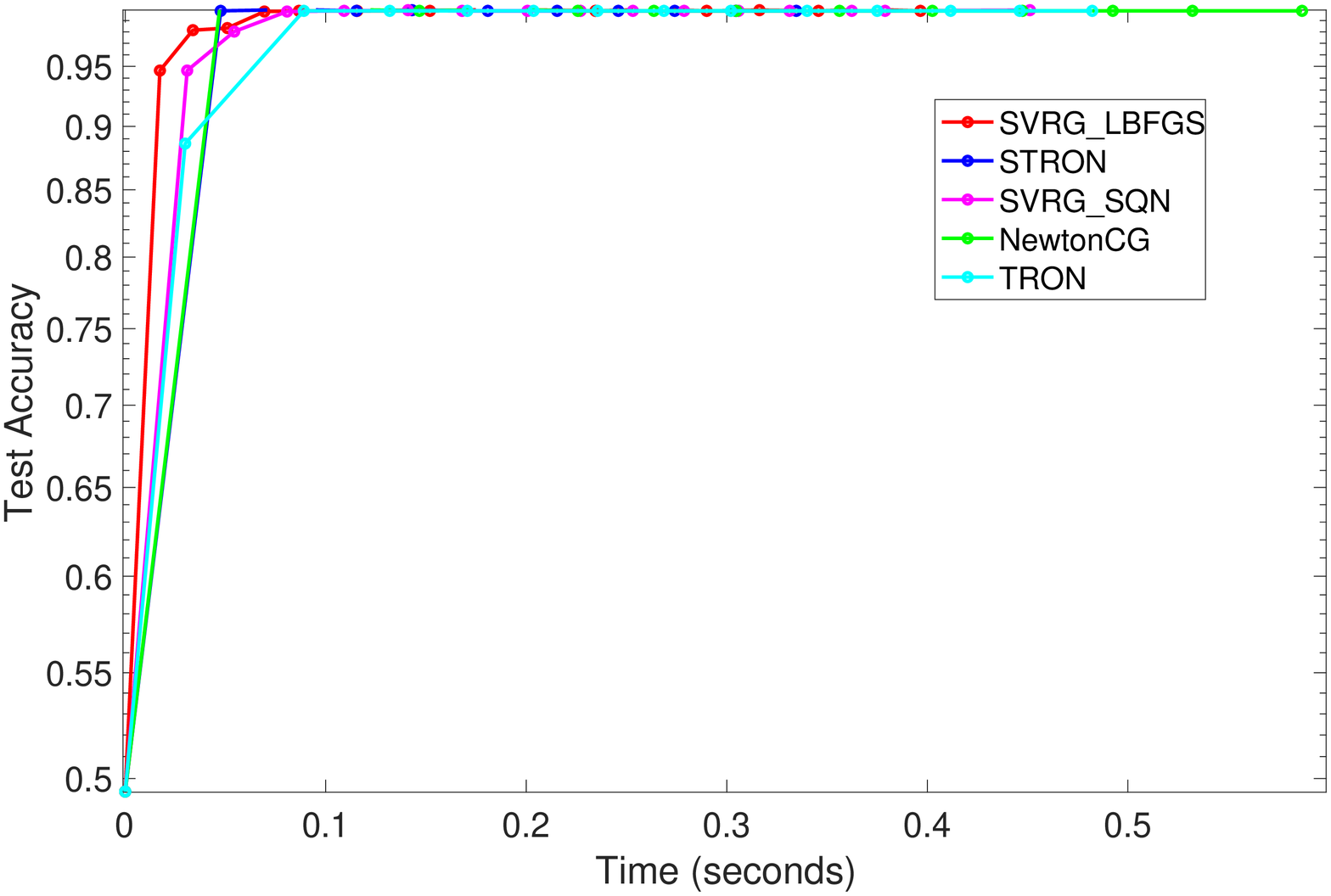}}
	
	\subfloat{\includegraphics[width=.5\linewidth]{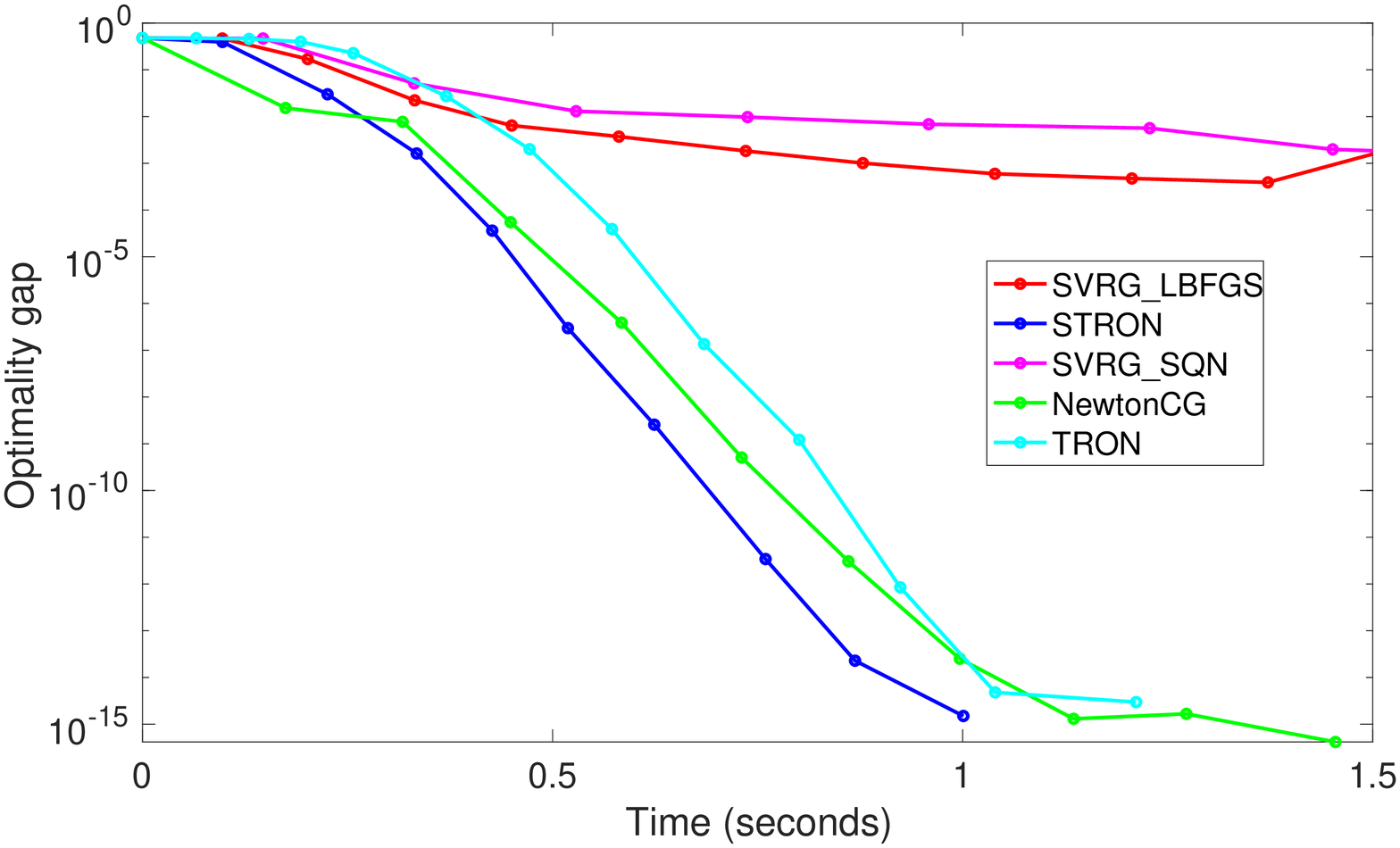}}
	\subfloat{\includegraphics[width=.5\linewidth]{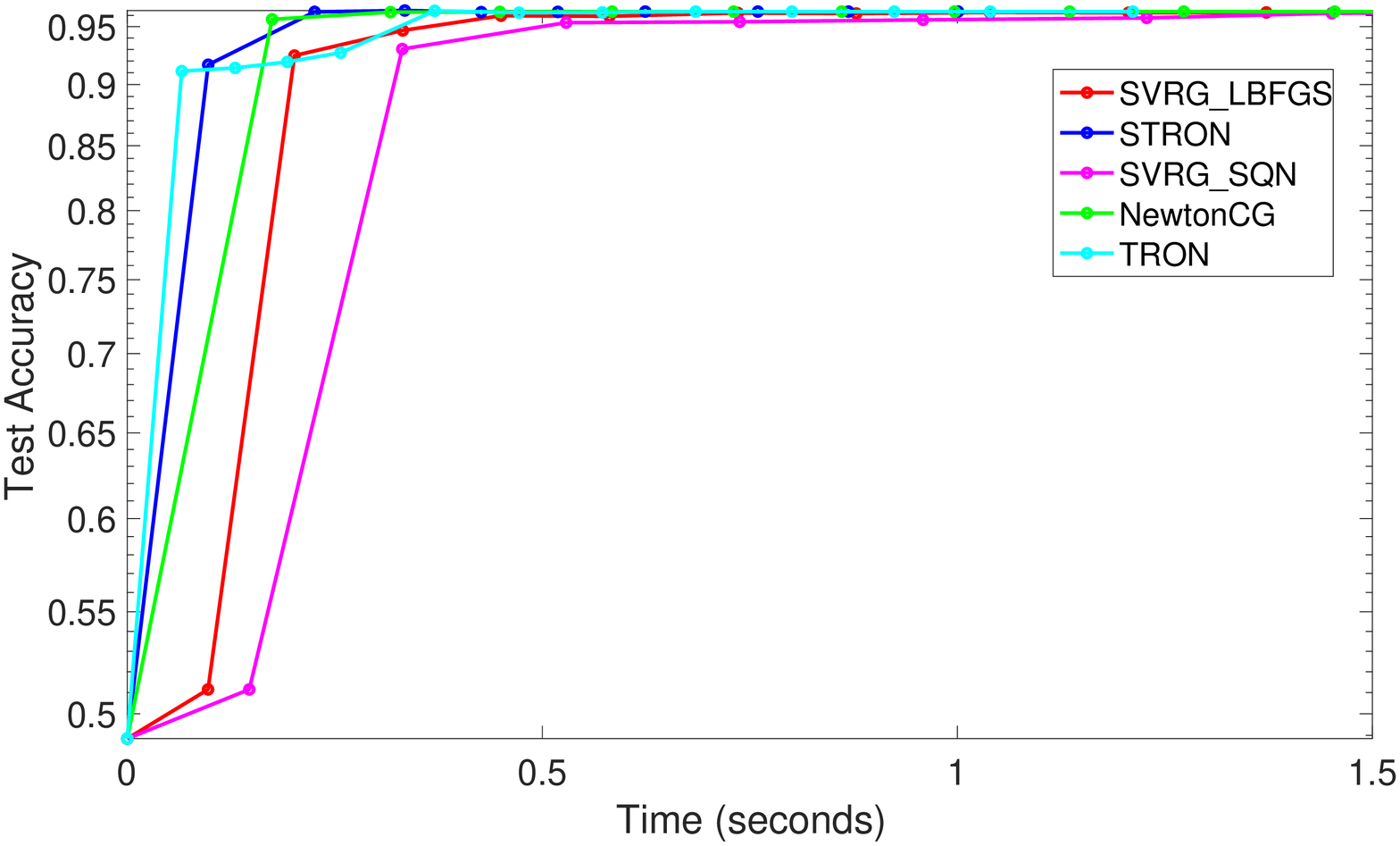}}
	
	\subfloat{\includegraphics[width=.5\linewidth]{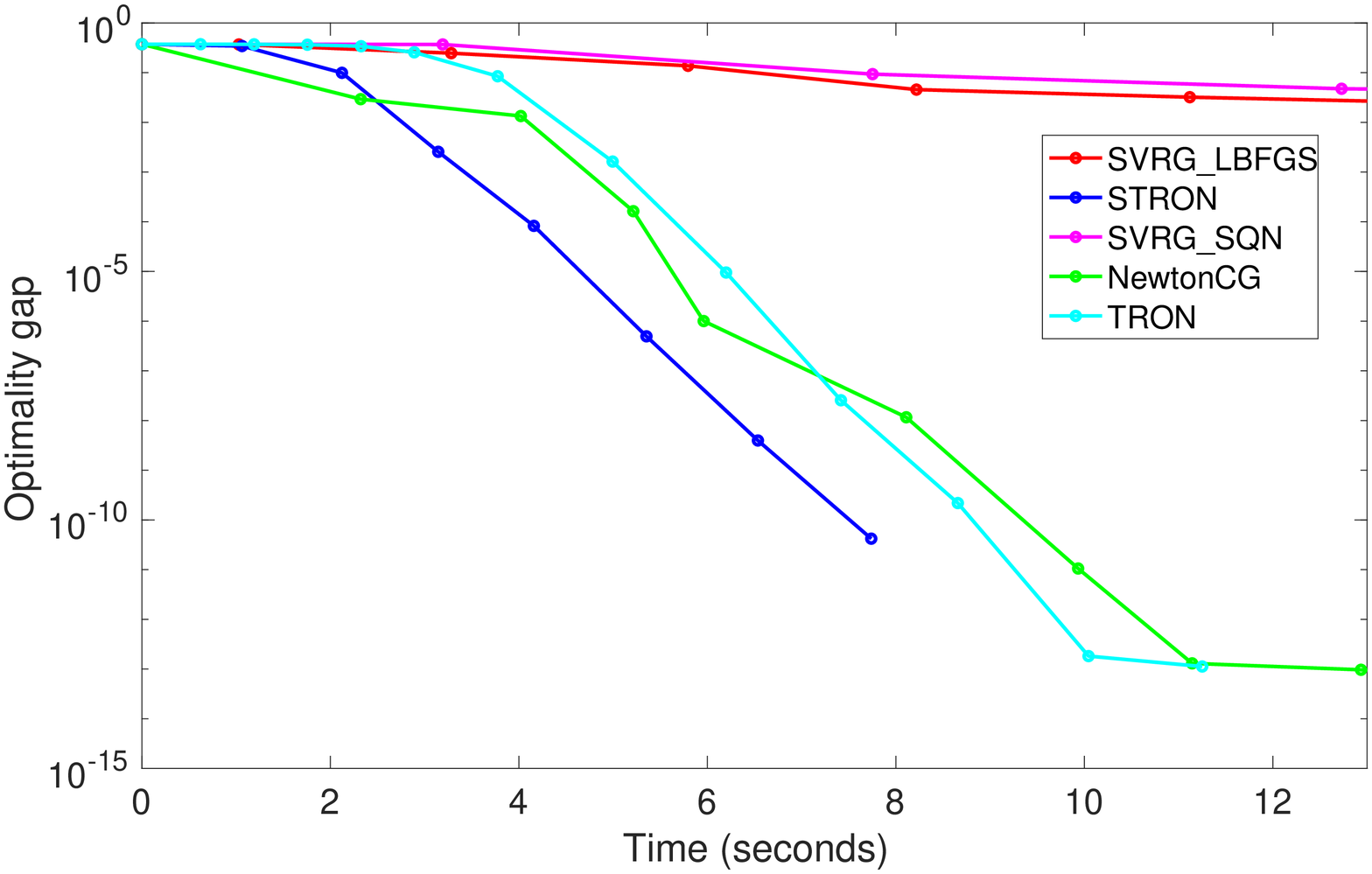}}
	\subfloat{\includegraphics[width=.5\linewidth]{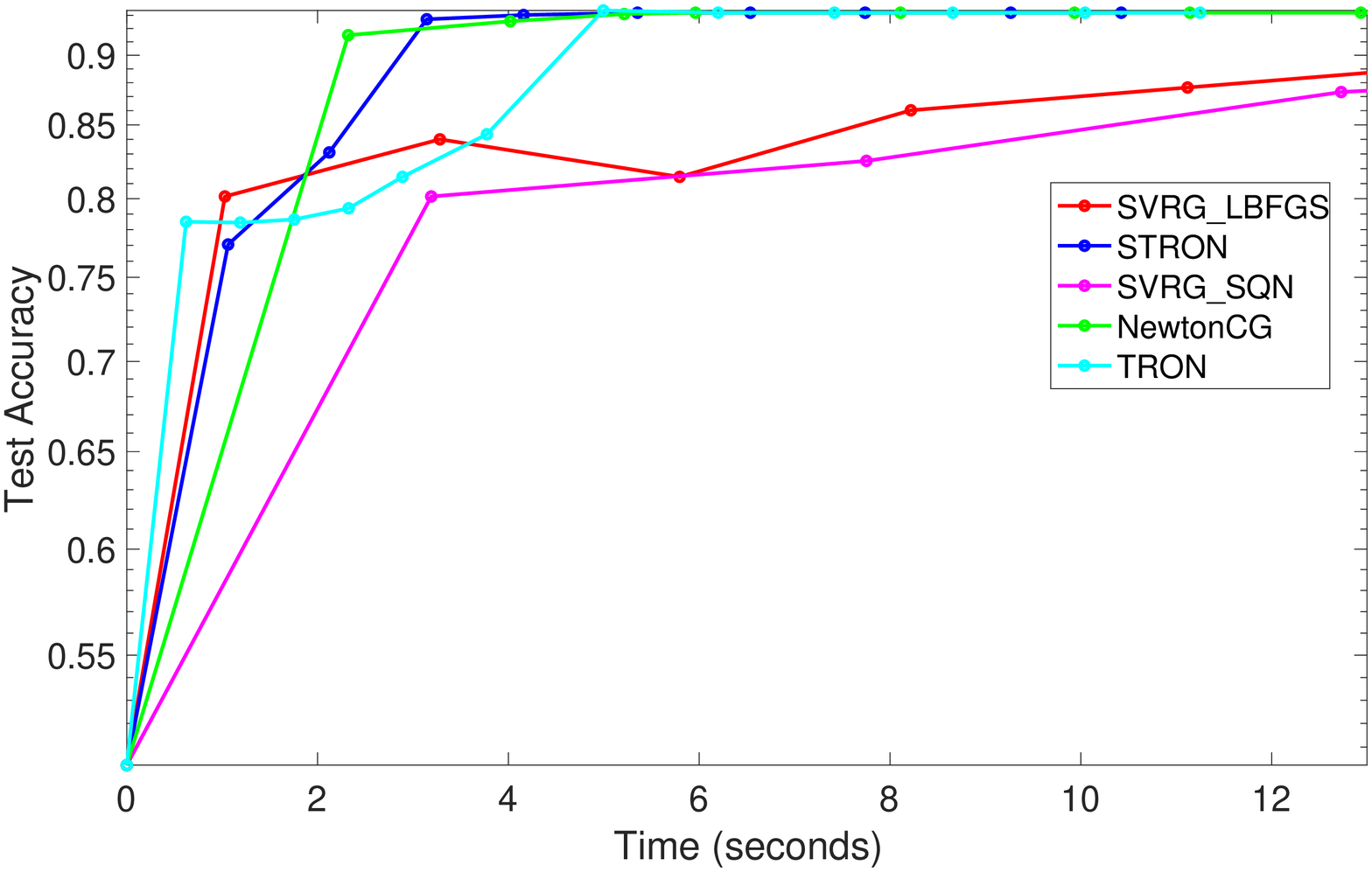}}
	
	\caption{First column presents optimality versus training time (in seconds) and second column presents accuracy versus training time, on mushroom (first row), rcv1 (second row) and news20 (third row) datasets.}
	\label{fig_1}
\end{figure}

\begin{figure}[htb]
	\subfloat{\includegraphics[width=.5\linewidth]{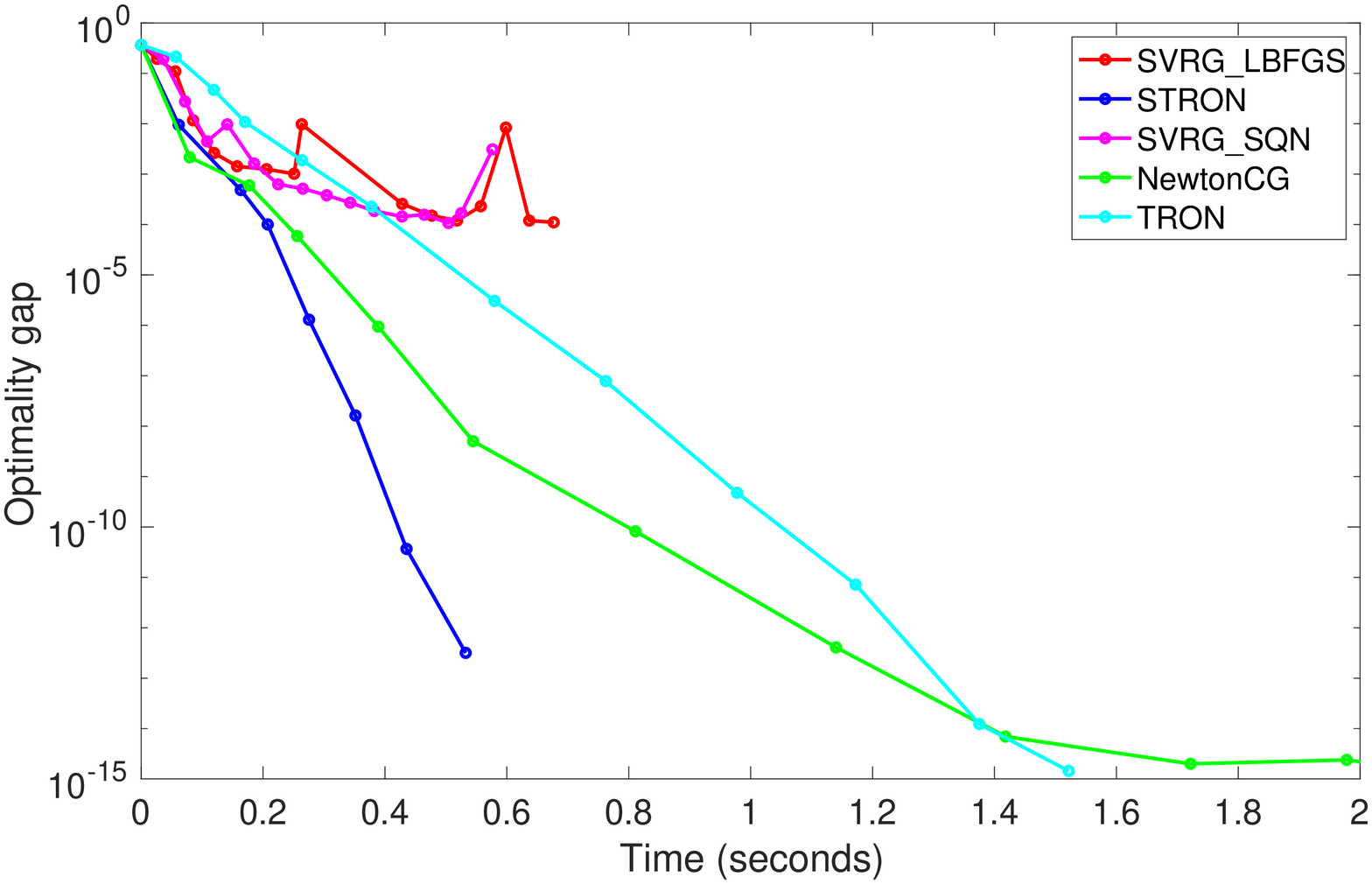}}
	\subfloat{\includegraphics[width=.5\linewidth]{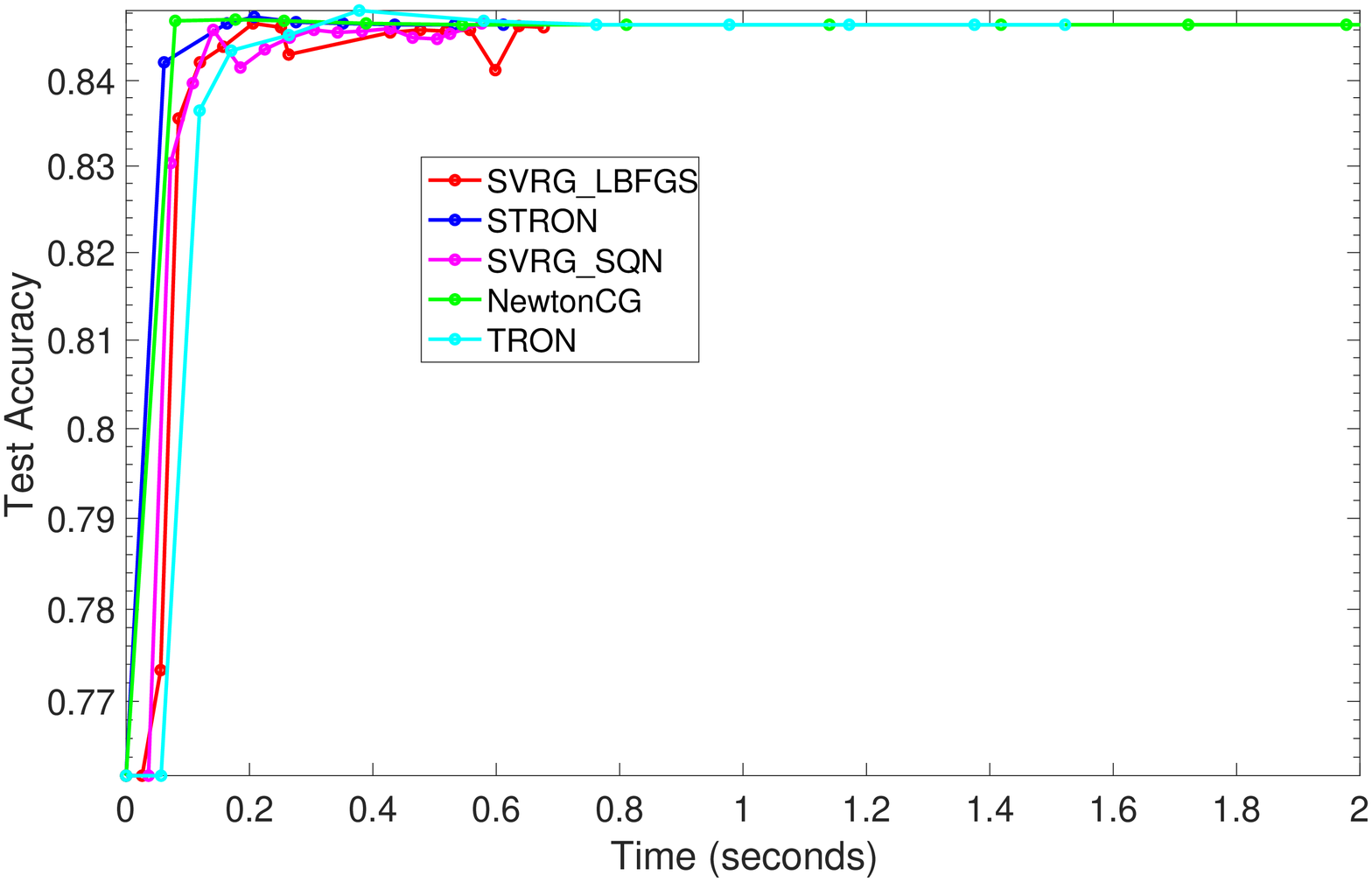}}
	
	\subfloat{\includegraphics[width=.5\linewidth]{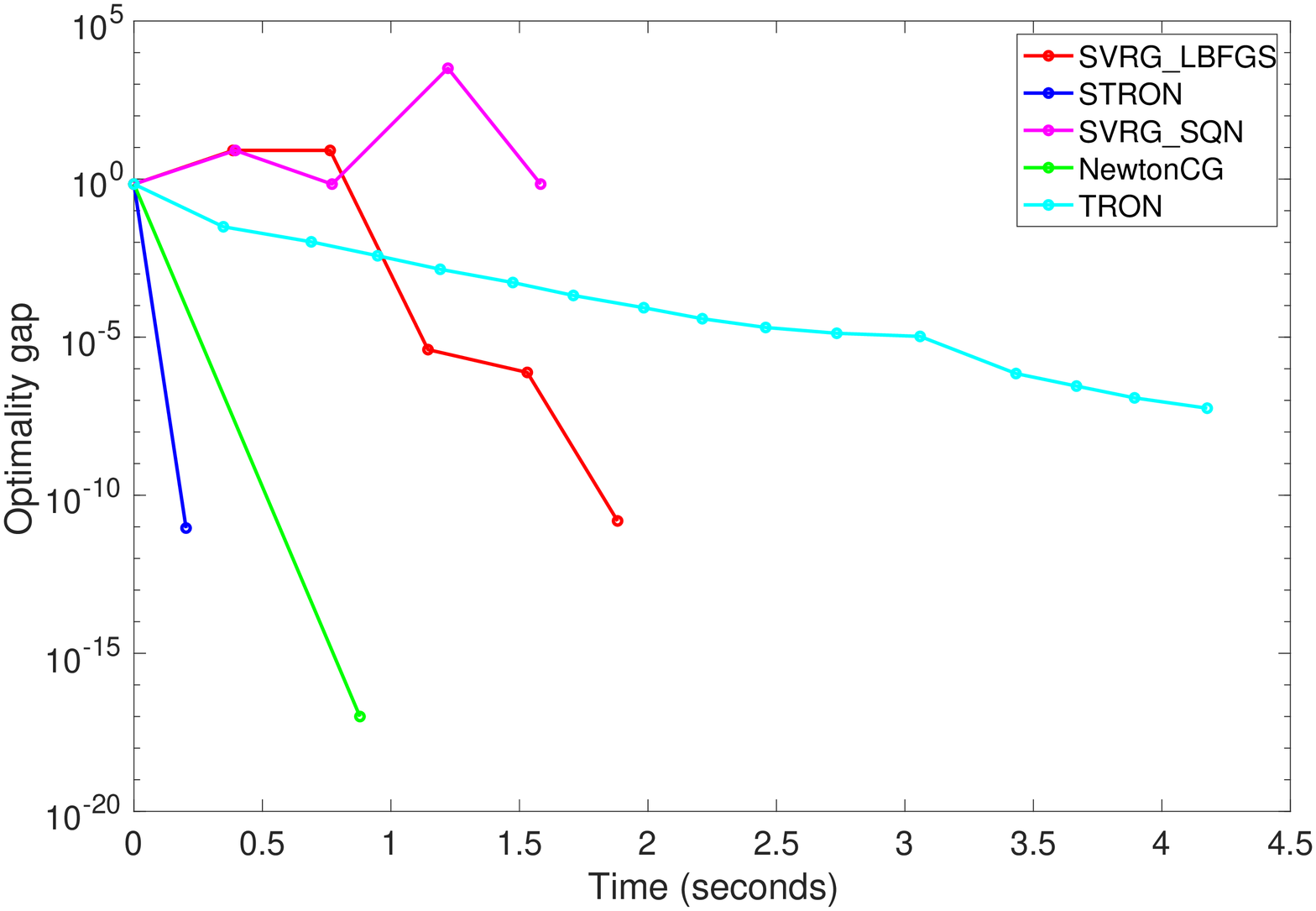}}
	\subfloat{\includegraphics[width=.5\linewidth,height=3.8cm]{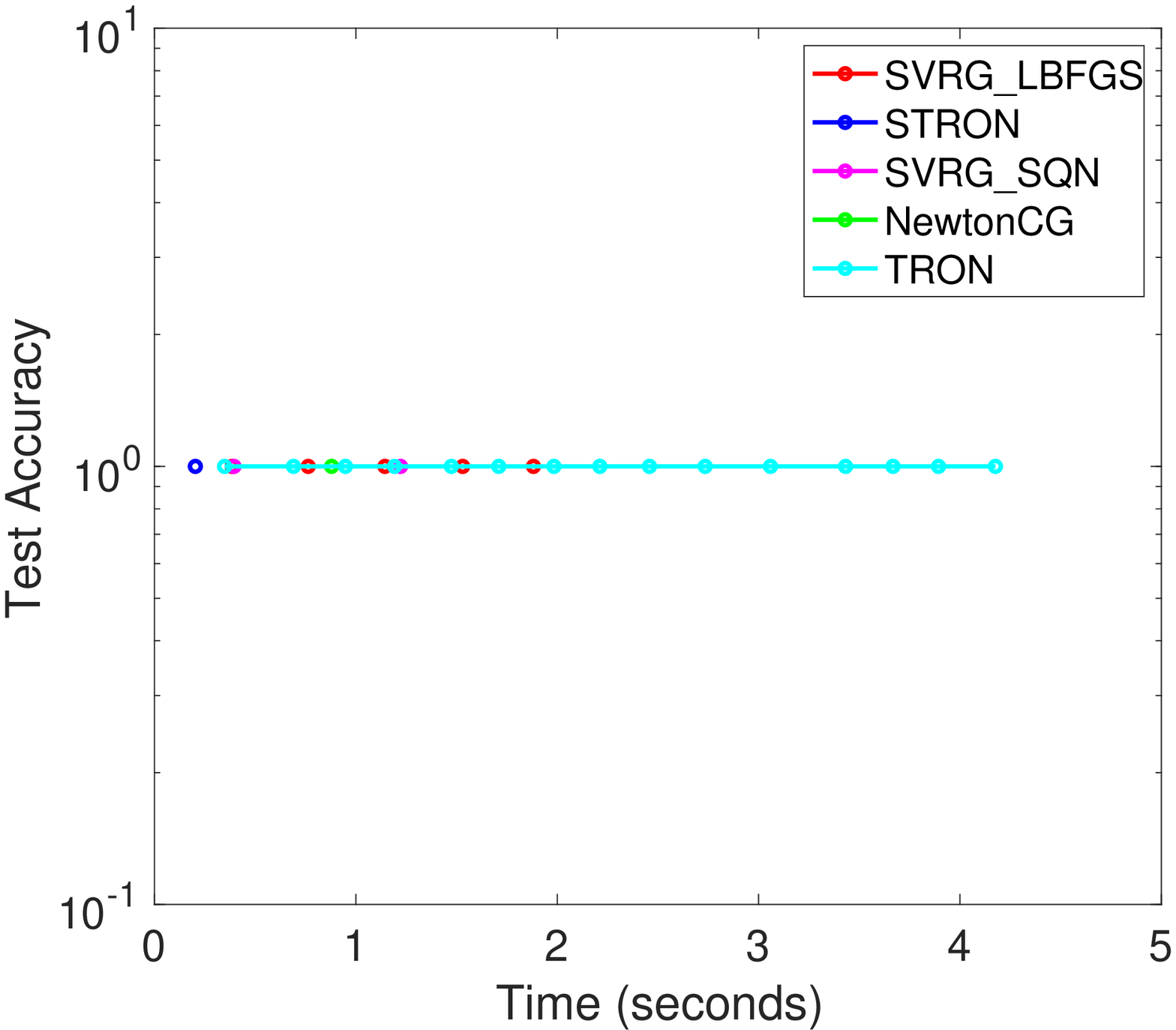}}
	
	\subfloat{\includegraphics[width=.5\linewidth]{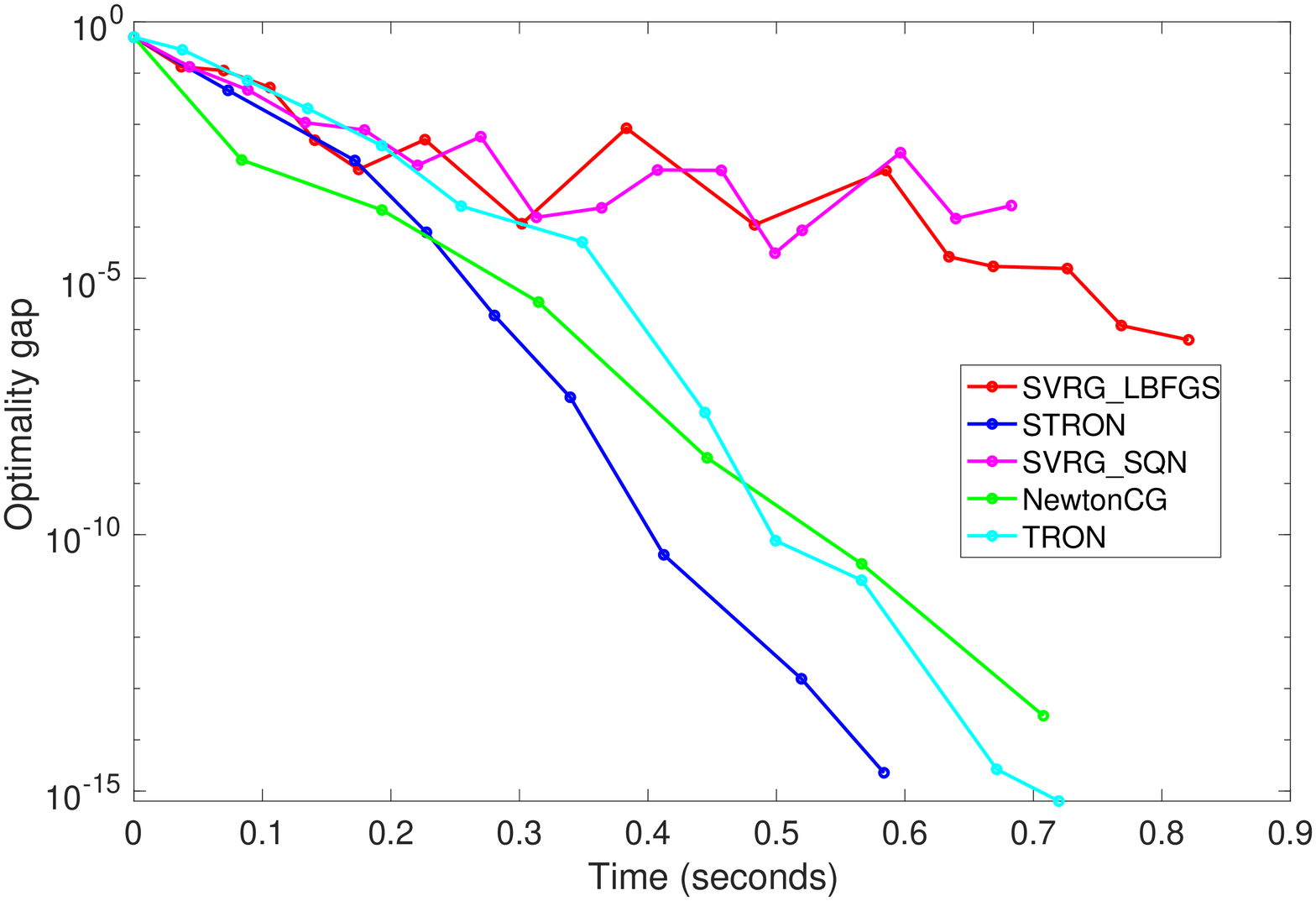}}
	\subfloat{\includegraphics[width=.5\linewidth]{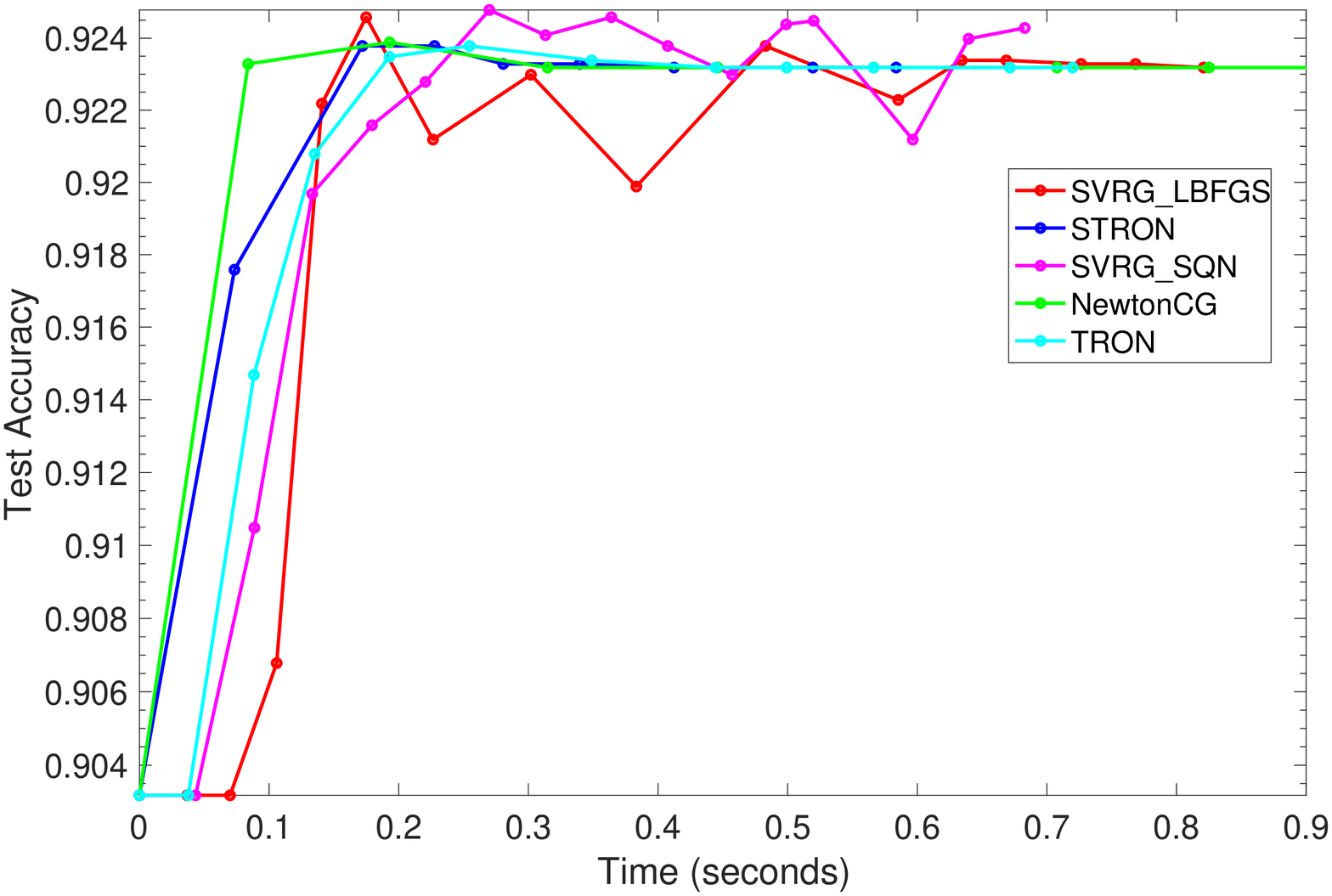}}
	
	\caption{First column presents optimality versus training time (in seconds) and second column presents accuracy versus training time, on Adult (first row), covtype (second row) and ijcnn1 (third row).}
	\label{fig_2}
\end{figure}

\begin{figure}[htb]
	\subfloat{\includegraphics[width=.5\linewidth]{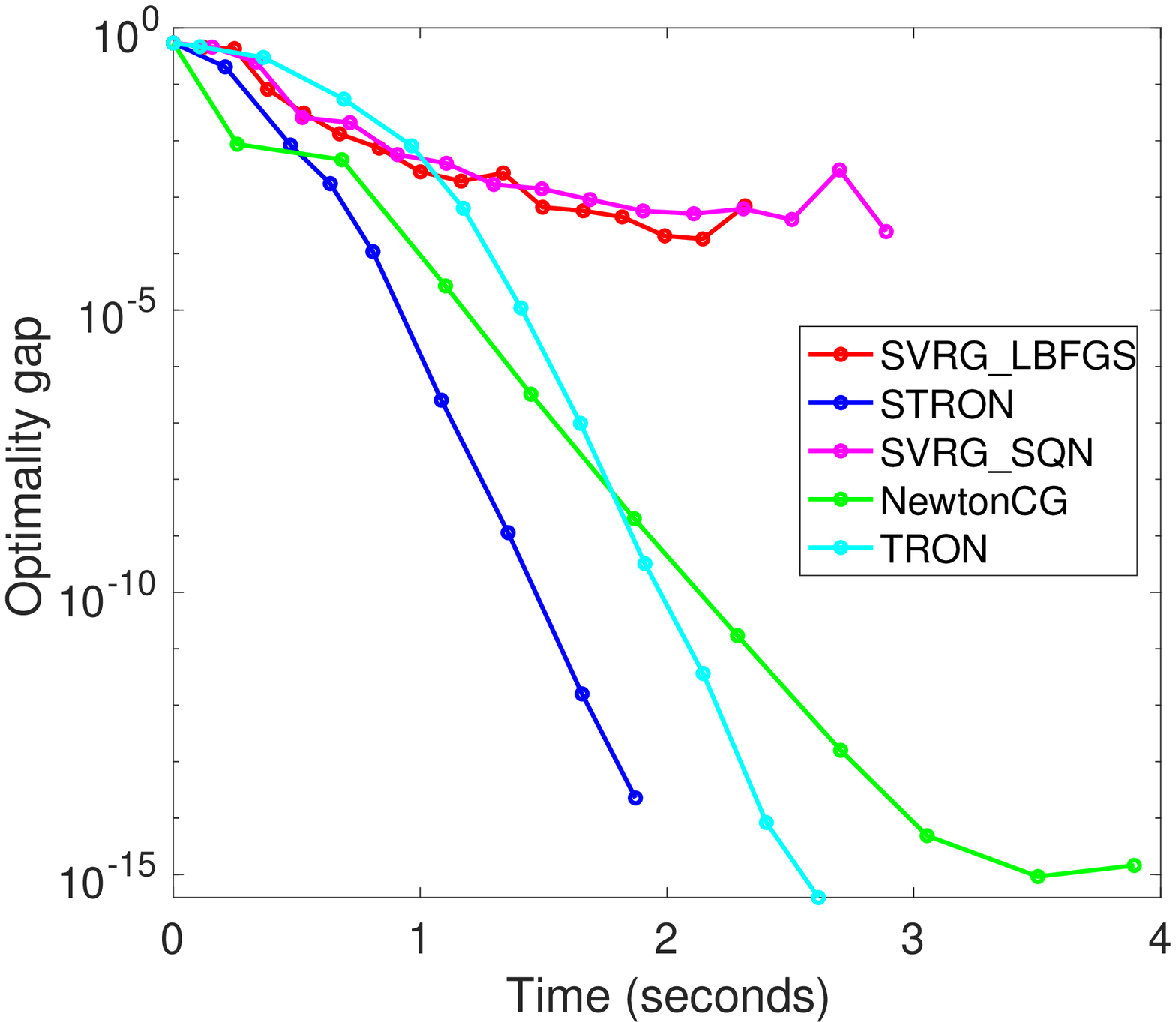}}
	\subfloat{\includegraphics[width=.5\linewidth]{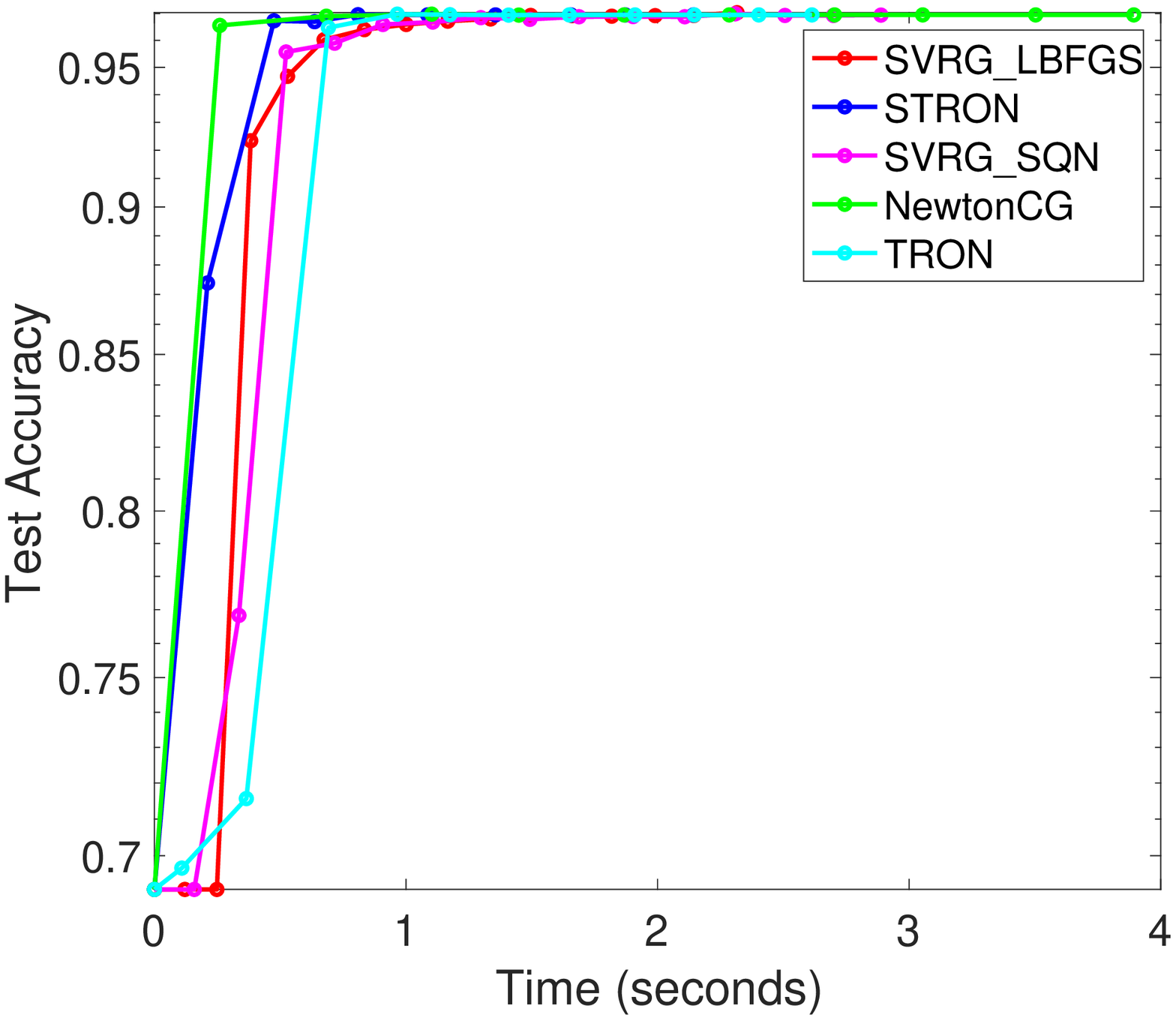}}
	
	\caption{First column presents optimality versus training time (in seconds) and second column presents accuracy versus training time, on real-sim dataset.}
	\label{fig_3}
\end{figure}

The experiments have been performed with strongly convex and smooth l$_2$-regularized logistic regression problem as given below:\\
\begin{equation}
	\label{eq_l2lr}
	\underset{w}{\min} \; F(w) = \dfrac{1}{l} \sum_{i=1}^{l} \log\left( 1 + \exp\left( - y_i w^T x_i \right) \right) + \dfrac{\lambda}{2} \|w\|^2.
\end{equation}
The results have been plotted as optimality ($F(w)-F(w^{*})$) versus training time (in seconds) and accuracy versus training time for high $\epsilon$(=$10^{-10}$)-accuracy solutions, as given in the Figs.~\ref{fig_1},~\ref{fig_2}~and~\ref{fig_3}. As it is clear from the results, STRON converges faster than all other methods and shows improvement against TRON on accuracy vs. time plots. Moreover, quasi-Newton methods converges slower than inexact Newton methods as already established in the literature \cite{TRON}. As per the intuitions, STRON takes initial advantage over TRON due to subsampling and as it reaches the solution region the progressive batching scheme reaches the full batching scheme and converges with same rate as TRON. That's why, in most of the figures, we can observe STRON and TRON converging in parallel lines. Moreover, we observe a horizontal line for accuracy vs. time plot with covtype dataset because all methods give 100\% accuracy.
Generally, the models are trained for low $\epsilon$(=$10^{-02}$)-accuracy solutions. So we present the results for such a case using Table~\ref{tab_time}, which reports results using 5-fold cross validation. As it is clear from the table, STRON either outperforms other solvers or shows results pretty close to the best method.
\begin{table}[h]
	\centering
	\caption{Comparison of Training Time (seconds) for low accuracy ($\epsilon$=0.01) solution}
	\label{tab_time}
	\begin{tabular}{|l|l|r|r|r|r|r|}
		\hline
		Datasets$\downarrow$& Methods$\rightarrow$ & SVRG\_LBFGS & STRON & SVRG\_SQN & Newton\_CG & TRON \\
		\hline
		\multirow{2}{*}{covtype} & Time  & 0.8440$\pm$0.1770 & \textbf{0.0620$\pm$0.0045} & 0.4140$\pm$0.0397 & 0.9740$\pm$0.1907 & 0.7294$\pm$0.0162 \\
		& Accuracy & \textbf{1.00$\pm$0.0} & \textbf{1.00$\pm$0.0} & \textbf{1.00$\pm$0.0} & \textbf{1.00$\pm$0.0} & \textbf{1.00$\pm$0.0} \\
		\hline
		\multirow{2}{*}{real-sim} & Time  & 1.5920$\pm$0.5867 & 0.9340$\pm$0.1258 & 2.3480$\pm$0.5909 & \textbf{0.68$\pm$0.01} & 0.88$\pm$0.10 \\
		& Accuracy & 0.9670$\pm$0.0055 & 0.9697$\pm$0.0017 & 0.9670 $\pm$0.0029& 0.9688$\pm$0.0022 & \textbf{0.9698$\pm$0.0015} \\
		\hline
		\multirow{2}{*}{rcv1} & Time  & 2.1780$\pm$0.1813& \textbf{0.5700$\pm$0.0200} & 3.2840$\pm$0.0270 & 0.5860$\pm$0.0167 & 0.6589$\pm$0.0866 \\
		& Accuracy & \textbf{0.9641$\pm$0.0023} & 0.9640$\pm$0.0025 & 0.9637$\pm$0.0024 & 0.9640$\pm$0.0024 & 0.9639$\pm$0.0027 \\
		\hline
		\multirow{2}{*}{news20} & Time  & 42.8960$\pm$2.5993 & 6.9620$\pm$1.3534 & 84.0520$\pm$2.2915 & \textbf{6.764$\pm$0.0074} & 7.2470$\pm$0.3567 \\
		& Accuracy & 0.9333$\pm$0.0079 & 0.9337$\pm$0.0076 & \textbf{0.9339$\pm$0.0072} & 0.9338$\pm$0.0074& 0.9338$\pm$0.0077 \\
		\hline
		\multirow{2}{*}{mushroom} & Time  & 0.1540$\pm$0.01340 & 0.0820$\pm$0.0268 &0.1820 $\pm$0.01920 & \textbf{0.072$\pm$0.0045} & 0.1316$\pm$0.0169 \\
		& Accuracy & 0.9995$\pm$0.0007 & 0.9992$\pm$0.0005& 0.9986$\pm$0.0026 & \textbf{0.9996$\pm$0.0005} & 0.9995$\pm$0.0005 \\
		\hline
		\multirow{2}{*}{ijcnn1} & Time  & 0.3460$\pm$0.1390 & \textbf{0.2560$\pm$0.03780} & 0.3600$\pm$0.1158 & 0.3120$\pm$0.0795 & 0.2564$\pm$0.0185 \\
		& Accuracy & 0.9235$\pm$0.0022 & 0.9232$\pm$0.0021 & 0.9237$\pm$0.0018 & \textbf{0.9238$\pm$0.0025} & 0.9233$\pm$0.0185 \\
		\hline
		\multirow{2}{*}{Adult} & Time  & 0.3620$\pm$0.0981 & \textbf{0.2000$\pm$0.0187} & 0.3100$\pm$0.0644 & 0.2380$\pm$0.0084 & 0.2139$\pm$0.0046 \\
		& Accuracy & \textbf{0.8479$\pm$0.0028} & 0.8470$\pm$0.0028 & 0.8469$\pm$0.0031 & 0.8473$\pm$0.0030 & 0.8477$\pm$0.0028 \\
		\hline
		\multirow{2}{*}{gisette} & Time  & 9.5620$\pm$0.9891 & \textbf{8.1800$\pm$0.4260} & 10.7020$\pm$1.1314 & 41.6480$\pm$36.1190 & 10.4277$\pm$0.1709 \\
		& Accuracy & 0.9725$\pm$0.0036 & 0.9730$\pm$0.0045 & 0.9690$\pm$0.0026 & 0.9701$\pm$0.0057 & \textbf{0.9755$\pm$0.0028} \\
		\hline
		\multirow{2}{*}{webspam} & Time  & 6.0480$\pm$2.2814 & 6.3720$\pm$0.9942 & 7.6080$\pm$2.1462 & \textbf{3.272$\pm$0.2141} & 9.0548$\pm$1.2572 \\
		& Accuracy & 0.9139$\pm$0.0033 & 0.9214$\pm$0.0019 & 0.9158$\pm$0.0036 & 0.9219$\pm$0.0017 & \textbf{0.9227$\pm$0.0009} \\
		\hline
	\end{tabular}
\end{table}

\subsection{Results with SVM}
\label{subsec_exp_svm}
We extend STRON to solve l$_2$-SVM problem which is a non smooth problem, as given below:
\begin{equation}
	\label{eq_l2svm}
	\underset{w}{\min} \; F(w) = \dfrac{1}{l} \sum_{i=1}^{l} \max(0, 1 - y_i w^T x_i)^2 + \dfrac{\lambda}{2} \|w\|^2.
\end{equation}
\begin{figure}[htb]
	\subfloat{\includegraphics[width=.5\linewidth]{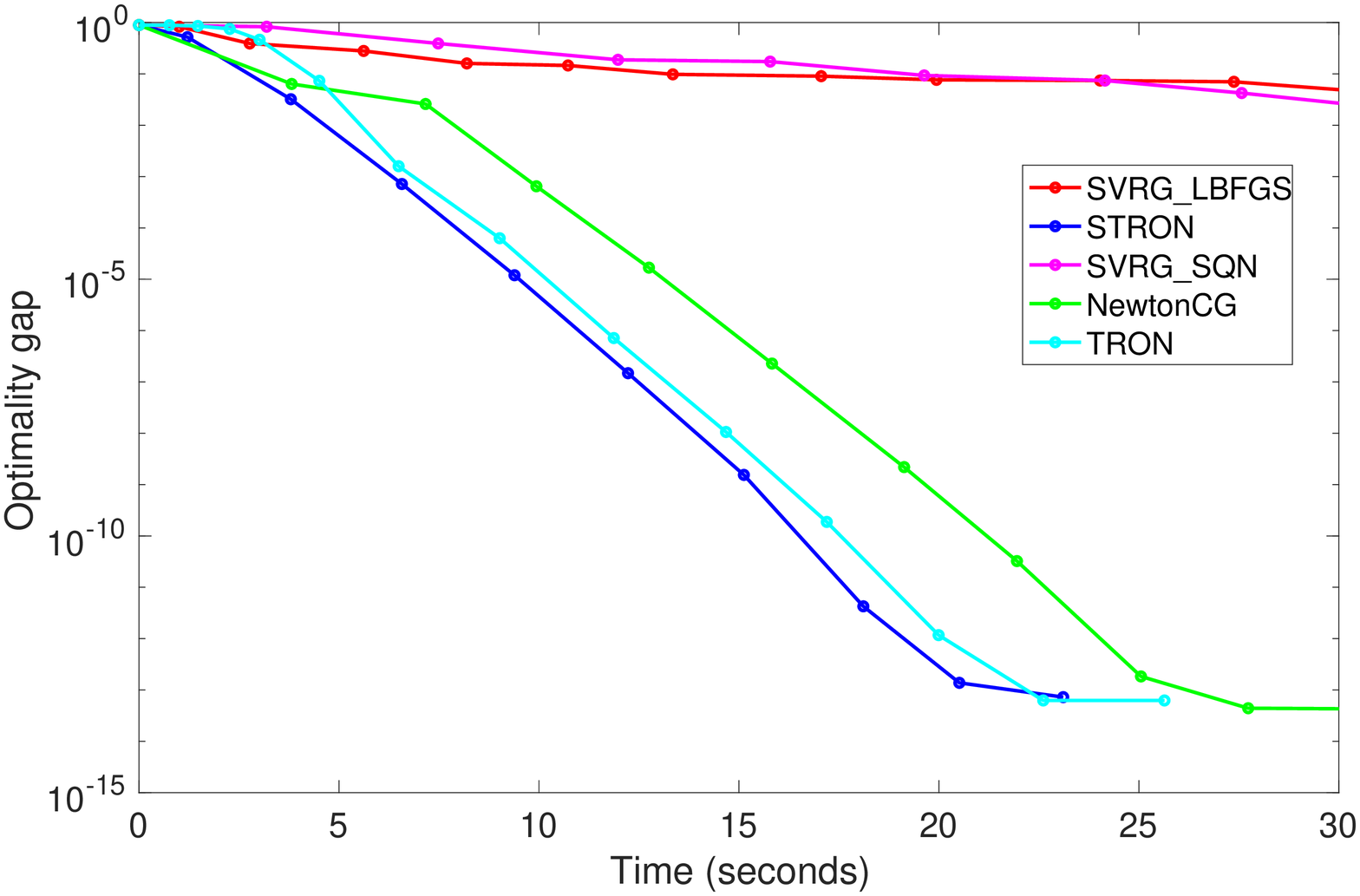}}
	\subfloat{\includegraphics[width=.5\linewidth]{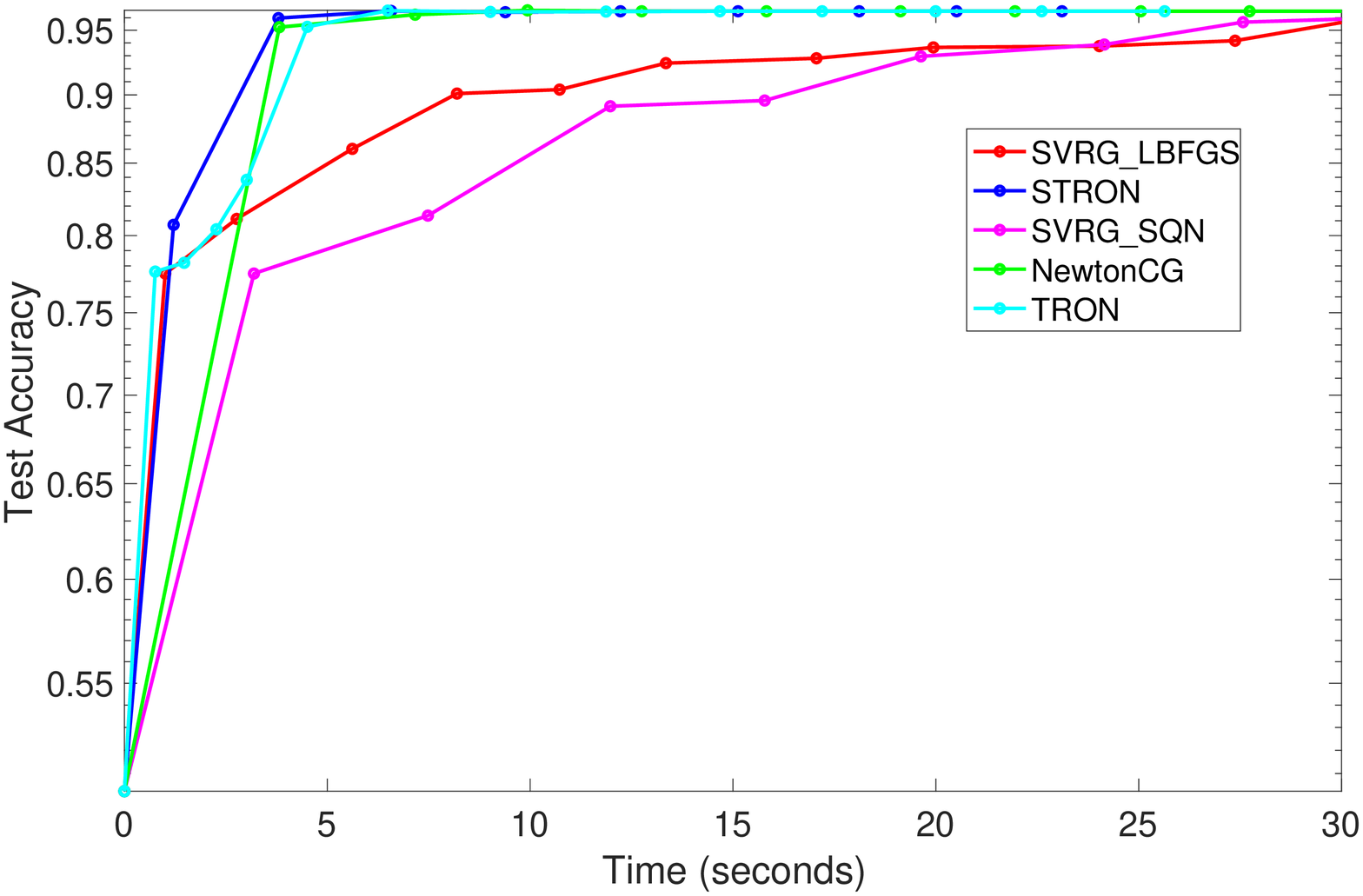}}
	
	\subfloat{\includegraphics[width=.5\linewidth]{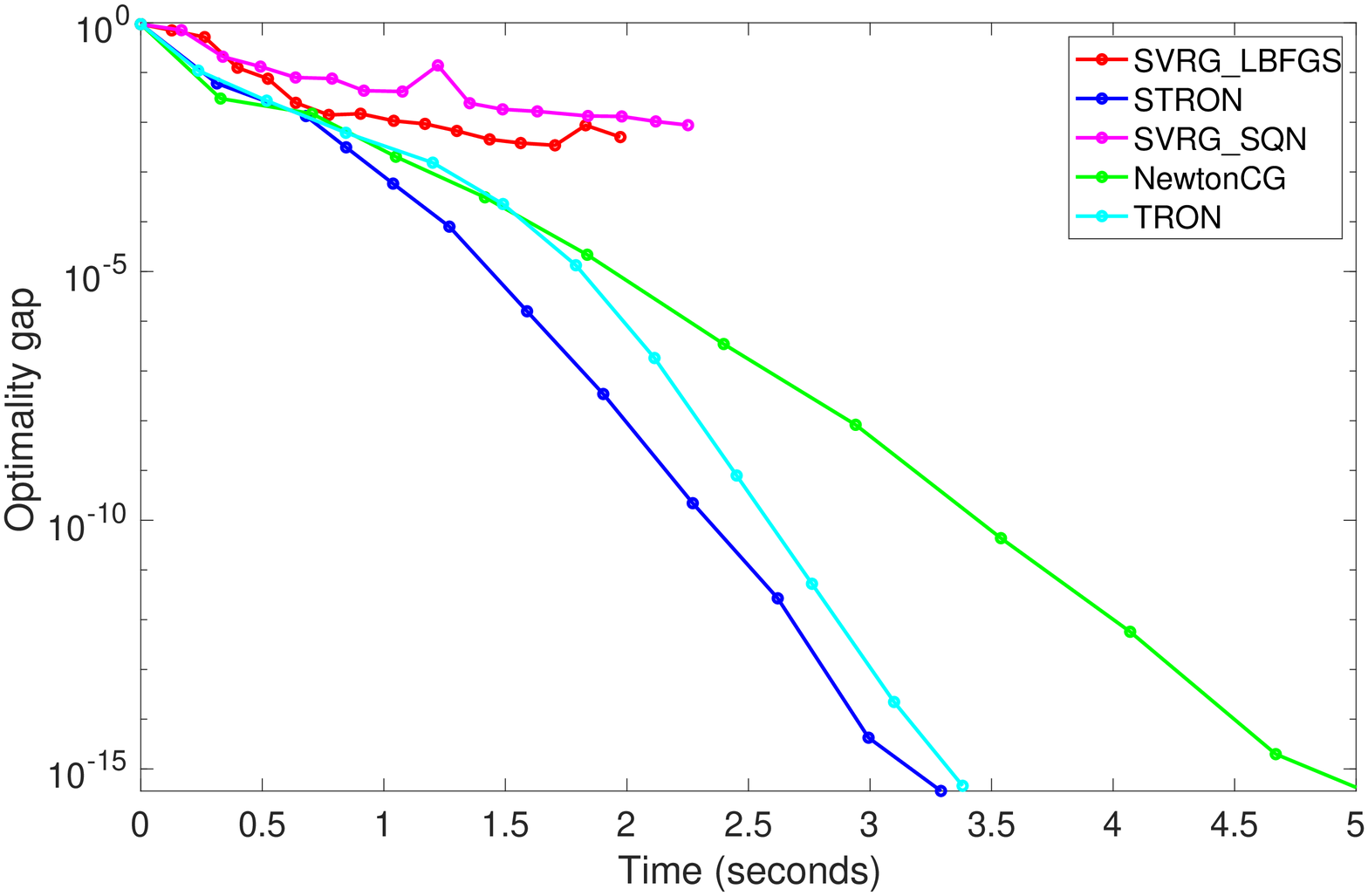}}
	\subfloat{\includegraphics[width=.5\linewidth]{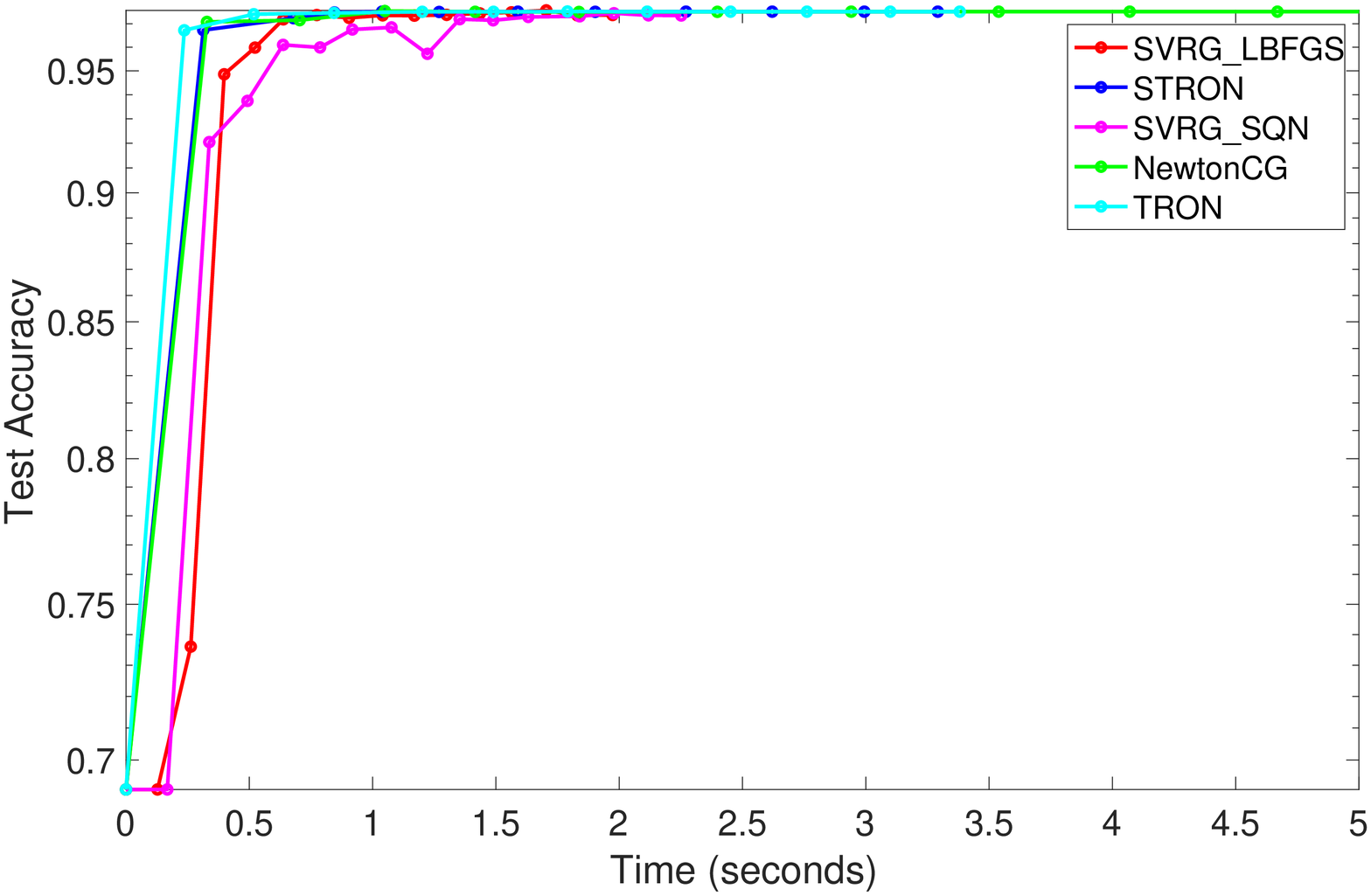}}
	
	\caption{Experiments with l$_2$-SVM on news20 (first row) and real-sim (second row) datasets.}
	\label{fig_svm}
\end{figure}
The results are reported in the Fig.~\ref{fig_svm} with news20 and real-sim datasets. As it is clear from the figure, STRON shows results similar to logistic regression problem and outperforms all other methods.

\section{Conclusion}
\label{discussion}
We proposed a novel stochastic trust region inexact Newton method, called as STRON, to solve the large-scale learning problems. The proposed method used progressive batching scheme to deal with noisy approximations of gradient and Hessian, and enjoyed the benefits of both stochastic and full batch regimes. STRON has been extended to use preconditioned CG as trust region subproblem solver and to use variance reduction for noisy gradient calculations. Our empirical results proved the efficacy of STRON against the state-of-art techniques with bench marked datasets.

\begin{acknowledgements}
First author is thankful to Ministry of Human Resource Development, Government of INDIA, to provide fellowship (University Grants Commission - Senior Research Fellowship) to pursue his PhD. We are also thankful to the anonymous reviewers for their constructive comments to improve the quality of our manuscript.
\end{acknowledgements}

% Authors must disclose all relationships or interests that 
% could have direct or potential influence or impart bias on 
% the work: 
%
 \section*{Conflict of interest}

 The authors declare that they have no conflict of interest.

\appendix

\section{Extensions}
\label{subsec_extensions}
In this section, we discuss extensions of the proposed method with PCG for solving the trust region subproblem, and with variance reduction technique.
\subsection{PCG Subproblem Solver}
\label{subsec_pcg}
\begin{figure}[htb]
	\subfloat{\includegraphics[width=.5\linewidth]{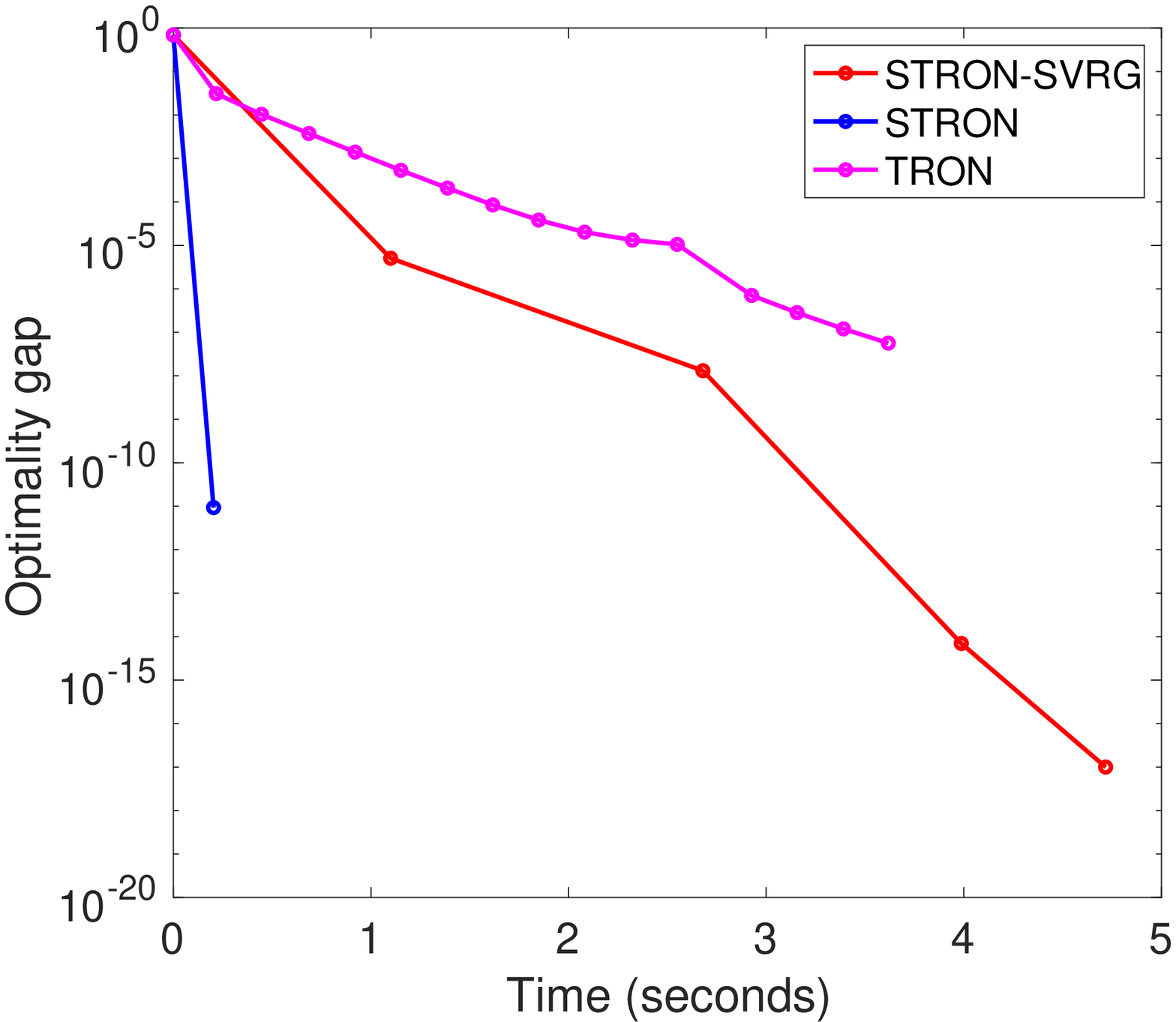}}
	\subfloat{\includegraphics[width=.5\linewidth]{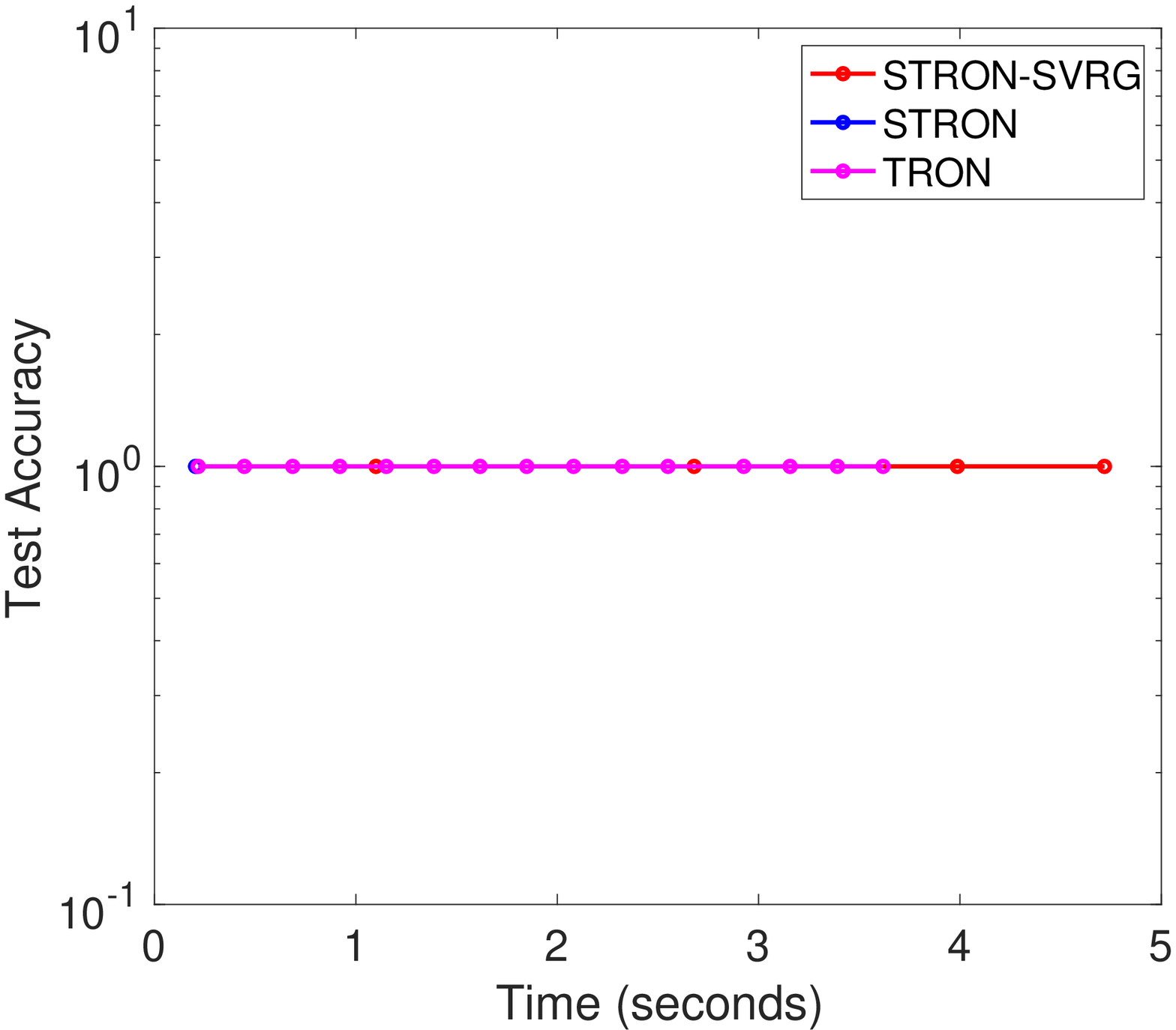}}
	
	\subfloat{\includegraphics[width=.5\linewidth]{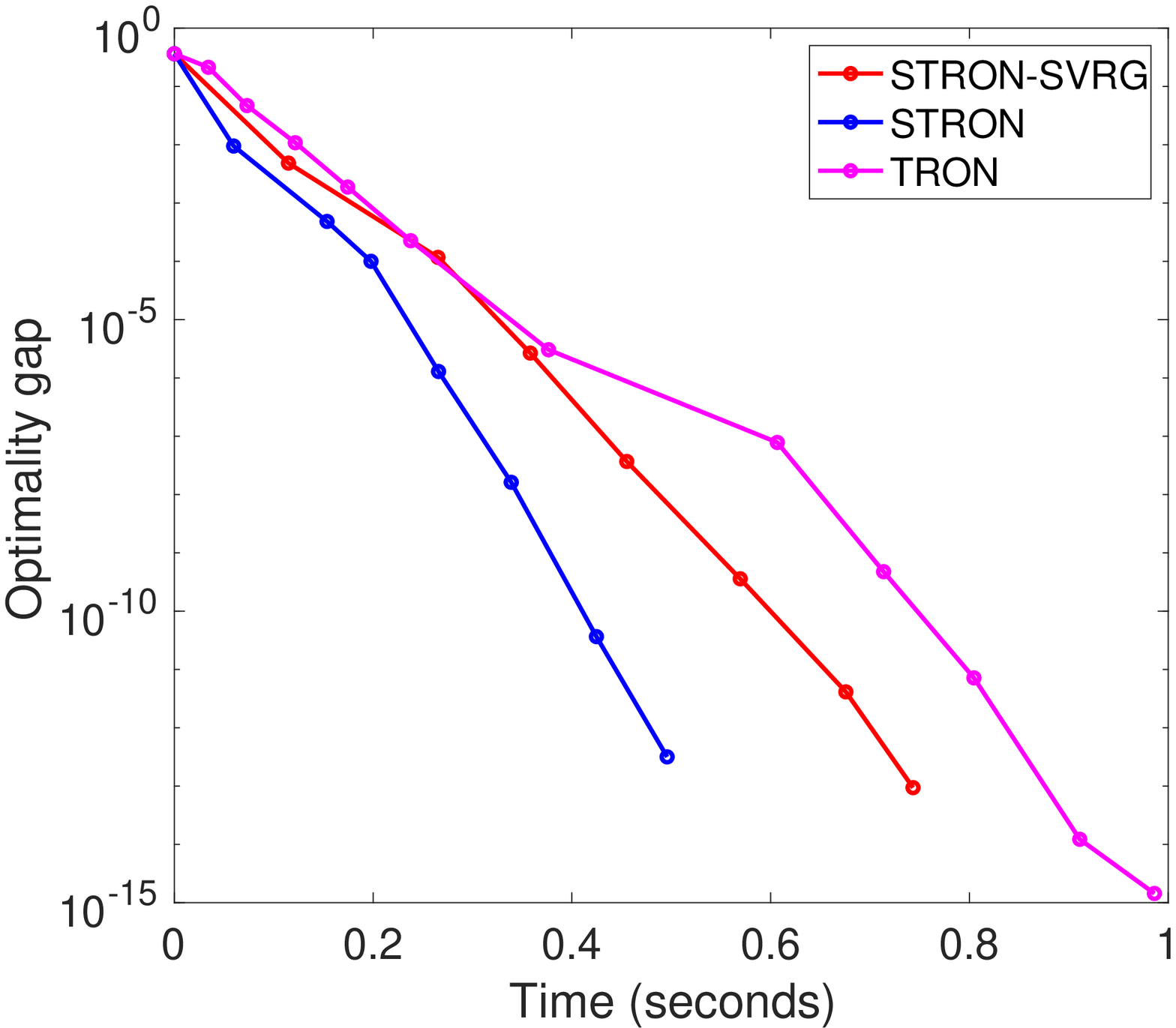}}
	\subfloat{\includegraphics[width=.5\linewidth]{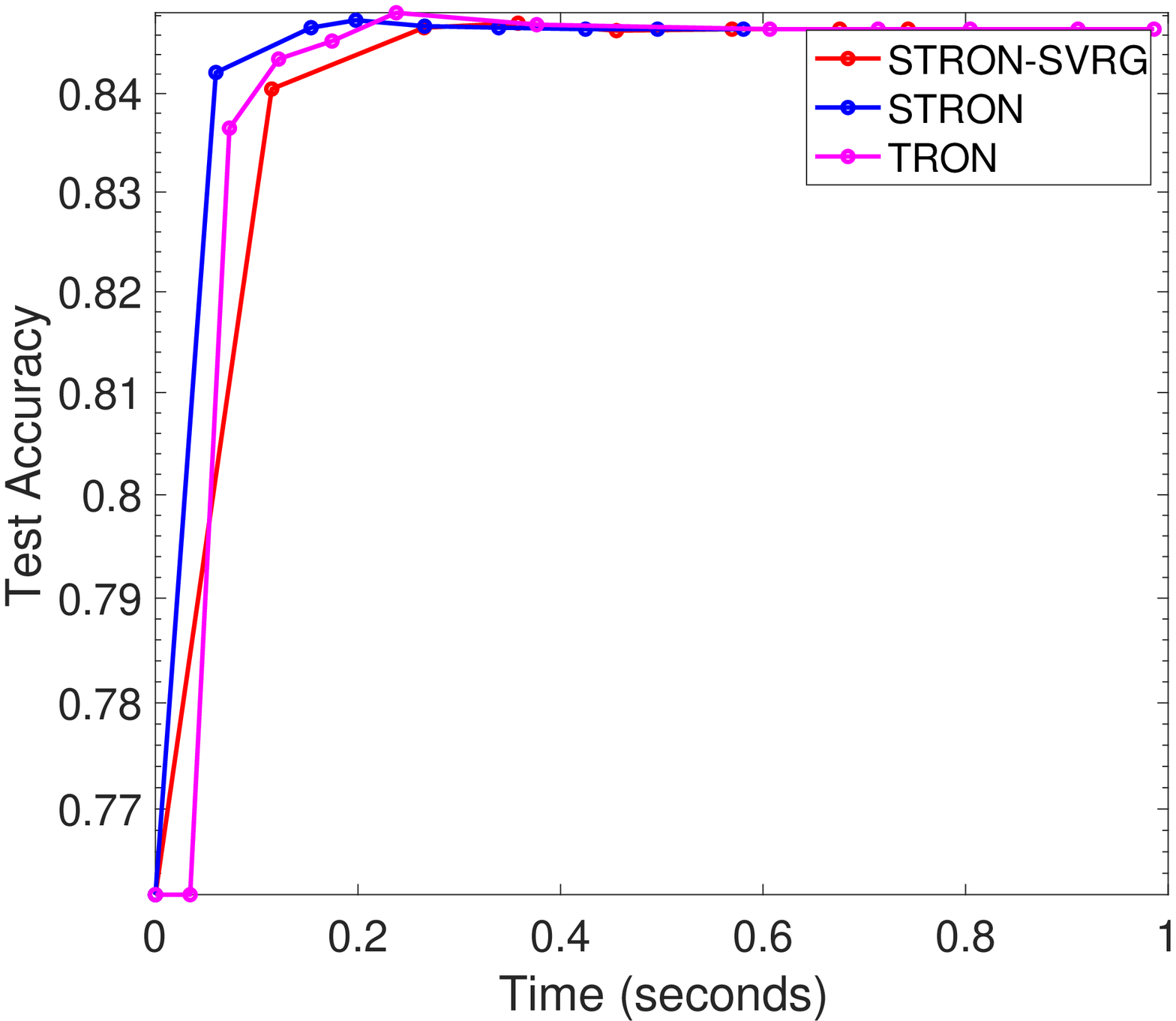}}
	
	\caption{Comparative study of STRON\_PCG, STRON and TRON on covtype (first row) and Adult (second row) datasets.}
	\label{fig_pcg}
\end{figure}

Number of iterations required by CG method to solve the subproblem depend on the condition number of the Hessian matrix. So for ill-conditioned problems CG method converges slowly. To avoid such situations, generally a non-singular matrix $M$, called preconditioner, is used as follow. For the linear system $\nabla^2 F(w) p = - \nabla F(w)$, we solve following system:
\begin{equation}
\label{eq_linear_system}
M^{-1}\nabla^2 F(w) p = - M^{-1}\nabla F(w).
\end{equation}
Generally, $M^{-1} = LL^T$ is taken to ensure the symmetry and positive definiteness of $M^{-1}\nabla^2 F(w)$. PCG can be useful for solving the ill-conditioned problems but it involves extra computational overhead. We follow \cite{Hsia2018} to use PCG as a weighted average of identity matrix and diagonal matrix of Hessian, as given below:
\begin{equation}
\label{eq_preconditioner}
M = \alpha \times diag(H) + (1-\alpha) \times I,
\end{equation}
where $H$ is a Hessian matrix and $0\le \alpha \le 1$. For $\alpha=0$, there is no preconditioning and for $\alpha=1$ it is a diagonal preconditioner. In the experiments, we have taken $\alpha=0.01$ for TRON and STRON-PCG \cite{Hsia2018}. To apply PCG to trust region subproblem, we can use Algorithm~\ref{algo_cg} without any modifications, after changing the trust region subproblem (\ref{eq_tr_prob}), as given below \cite{Steihaug1983}:\\
\begin{equation}
\label{eq_pcg_subproblem}
\min_{\hat{p}} \left( L^{-1}\nabla F_{X_k}\left(w_k\right)\right)^T \hat{p} + \dfrac{1}{2} \hat{p}^T \left( L^{-1}\nabla ^2 F_{S_k}\left(w_k\right)L^{-T}\right) \hat{p}, \;\; \text{s.t.} \;\; \|\hat{p}\| \le \triangle_k,
\end{equation}
where $\hat{p} = L^{T} p$. STRON using PCG as a trust region subproblem solver is denoted by STRON-PCG and the results are reported in Fig.~\ref{fig_pcg}. It compares TRON, STRON and STRON-PCG on news20 and rcv1 datasets. As it is clear from the figure, both STRON and STRON-PCG outperform TRON.\\
PCG trust region subproblem solver involves extra cost for calculating the preconditioner, and for TRON the overhead due to preconditioner is given by
\begin{equation}
\label{eq_complexity_tron_pcg}
\begin{array}{ll}
& O\left(n\right) \times \#\text{CG iterations} + O(nl).
\end{array}
\end{equation}
And for STRON-PCG, preconditioner involves extra cost as given below:
\begin{equation}
\label{eq_complexity_stron_pcg}
\begin{array}{ll}
& O\left(n\right) \times \#\text{CG iterations} + O(n|S_k|).
\end{array}
\end{equation}

\subsection{Stochastic Variance Reduced Trust Region Inexact Newton Method}
\label{subsec_svrg}
\begin{figure}[htb]
	\subfloat{\includegraphics[width=.5\linewidth]{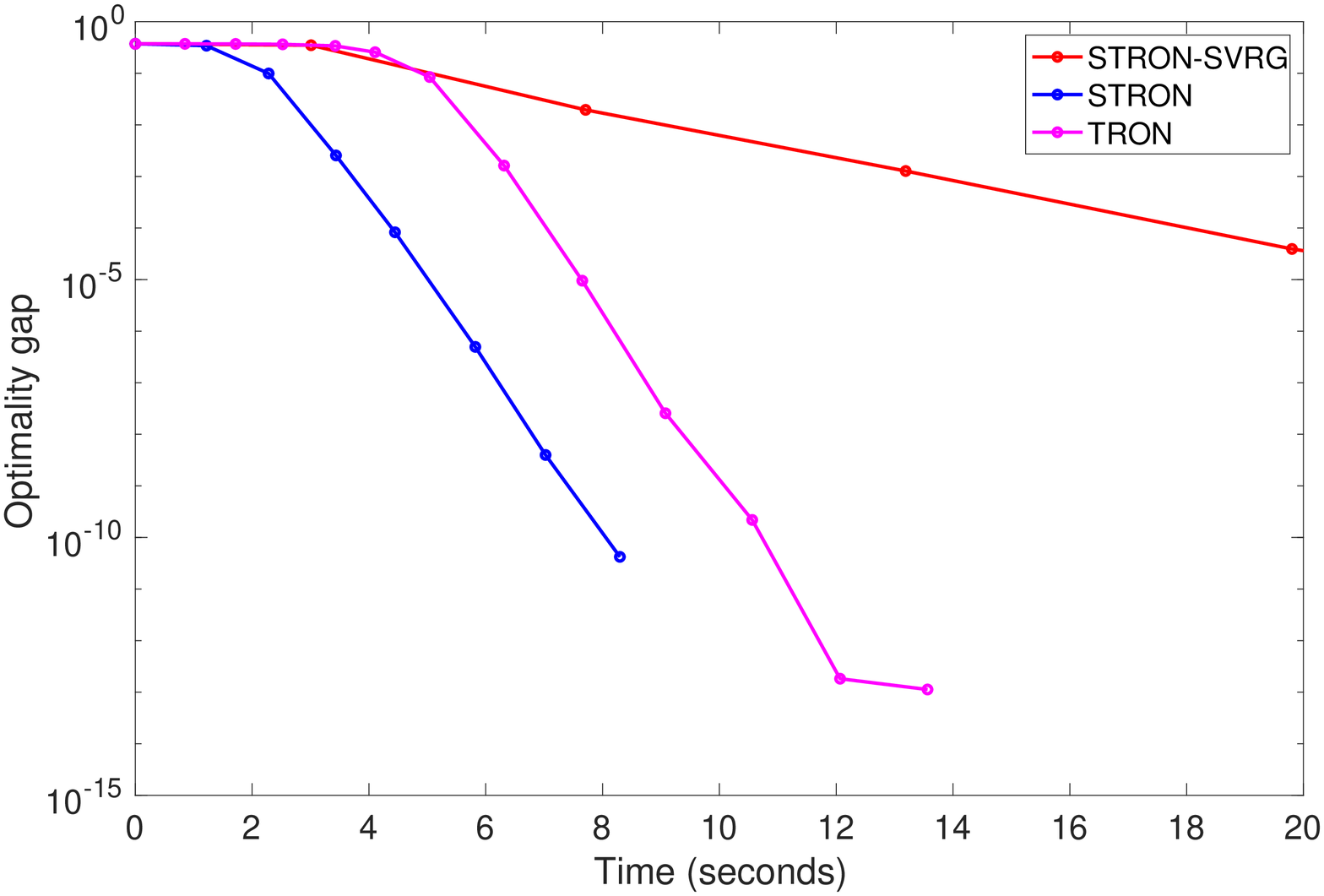}}
	\subfloat{\includegraphics[width=.5\linewidth]{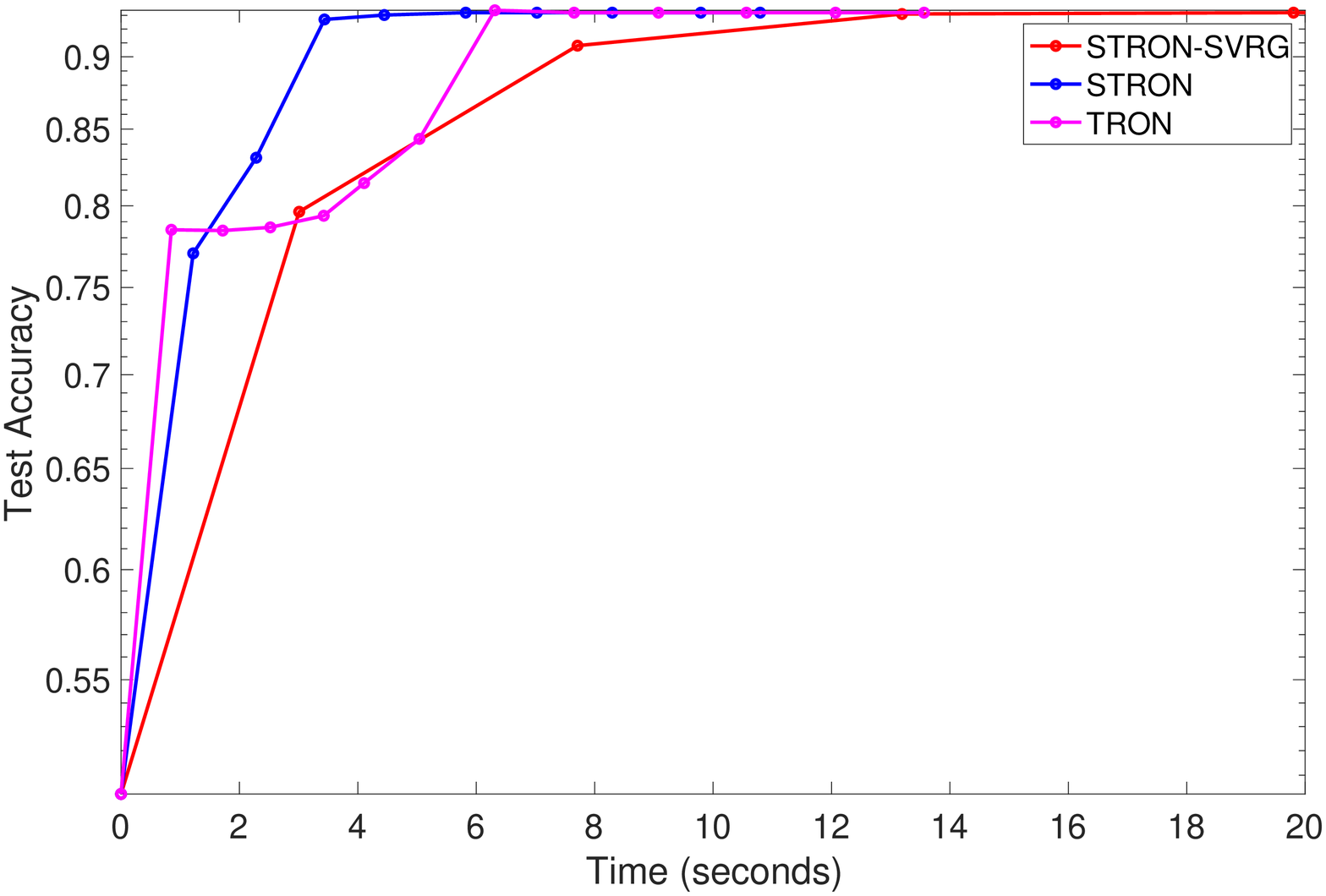}}
	
	\subfloat{\includegraphics[width=.5\linewidth]{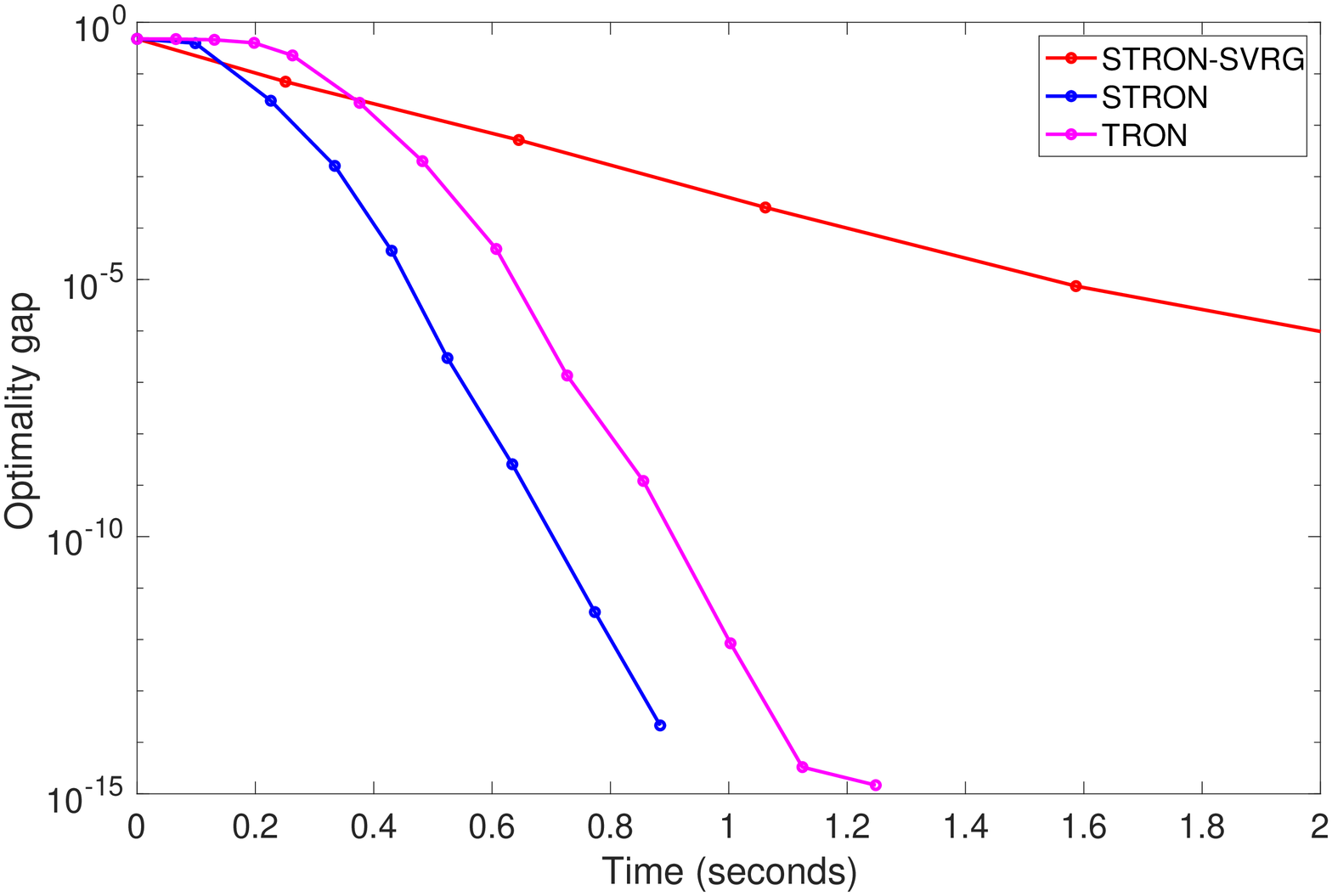}}
	\subfloat{\includegraphics[width=.5\linewidth]{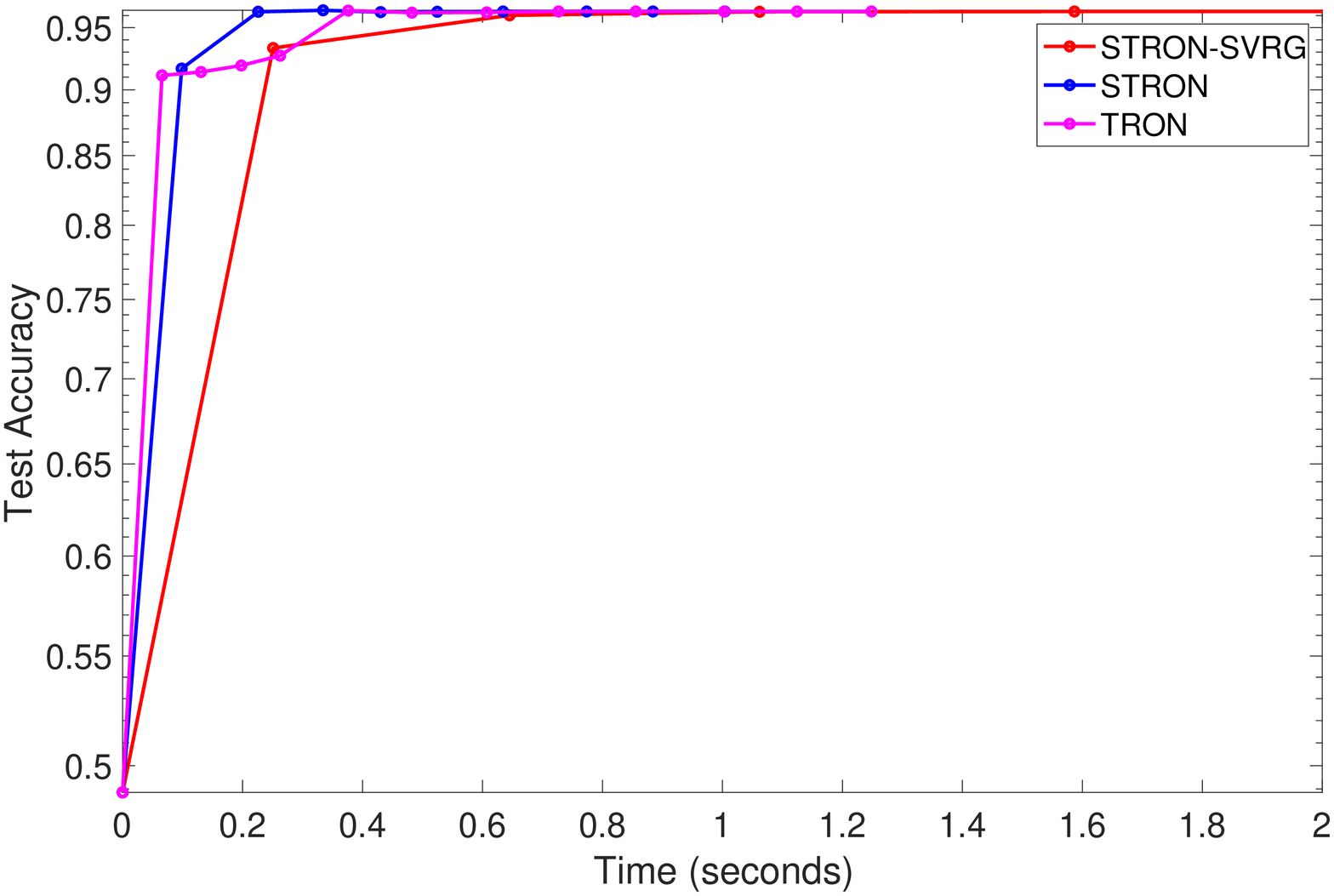}}
	
	\caption{Comparative study of STRON-SVRG, STRON and TRON on news20 (first row) and rcv1 (second row) datasets.}
	\label{fig_svrg}
\end{figure}

Recently, researchers have proposed stochastic variants of second order methods with variance reduction. So it is an interesting question to know that how will variance reduction work with stochastic trust region inexact Newton methods, as this is not studied yet. Our empirical results prove that variance reduction does not work in this case, even after using progressive subsampling for Hessian calculation.\\
\indent To improve the quality of search direction, we have used SVRG as variance reduction technique for gradient calculations, as given below:
\begin{equation}
\label{eq_svrg}
g_k = \nabla F_{S_k} (w_k) - \nabla F_{S_k} (\bar{w}) + \nabla F(\bar{w}),
\end{equation}
where $\bar{w}$ is parameter value at the start of outer iteration. STRON-SVRG uses variance reduction for gradient calculations and progressive batching for Hessian calculations, as given in the Algorithm~\ref{algo_stron_svrg}.
\begin{algorithm}[htb]
	\caption{STRON with Variance Reduction}
	\label{algo_stron_svrg}
	\begin{algorithmic}[1]
		\STATE \textbf{Inputs:} $w_0, m$
		\STATE \textbf{Result:} $w = w_k$
		\FOR{$i=0,1,2,...$}
		\STATE Calculate $\nabla F(w_i)$ and set $\bar{w} = w_i$
		\FOR{$k=0,1,...,(m-1)$}
		\STATE Randomly select subsamples $S_k$ and $X_k$
		\STATE Calculate subsampled gradient $\nabla F_{X_k}(w_k)$
		\STATE Calculate variance reduced gradient using (\ref{eq_svrg})
		\STATE Solve the trust region subproblem using Algorithm~\ref{algo_cg} with variance reduced gradient, instead of subsampled gradient, to get the step direction $p_k$
		\STATE Calculate the ratio $\rho_k = \left(F_{X_k}(w_k + p_k) - F_{X_k}(w_k)\right)/m_k(p_k)$
		\STATE Update the parameters using (\ref{eq_parameter_update})
		\STATE Update the trust region $\triangle_k$ using (\ref{eq_trust_region})
		\ENDFOR
		\ENDFOR
	\end{algorithmic}
\end{algorithm}
The experimental results are presented in Fig.~\ref{fig_svrg} with news20 and rcv1 datasets. As it is clear from the figures, STRON-SVRG lags behind STRON and TRON, i.e., variance reduction in STRON-SVRG is not sufficient to beat the progressive batching in gradient calculations of STRON. This is because both, STRON-SVRG and STRON, are stochastic/subsampled variants of TRON and to compensate for the noisy gradient calculations, former uses well-known variance reduction strategy but later uses progressive subsampling strategy. STRON is able to beat TRON but STRON-SVRG fails and lags behind both.

\section{More Results}
In this appendix, we provide more results and study the effect of regularization coefficient on the methods.
\subsection{More Experiments}
\begin{figure}[htb]
	\subfloat{\includegraphics[width=.5\linewidth]{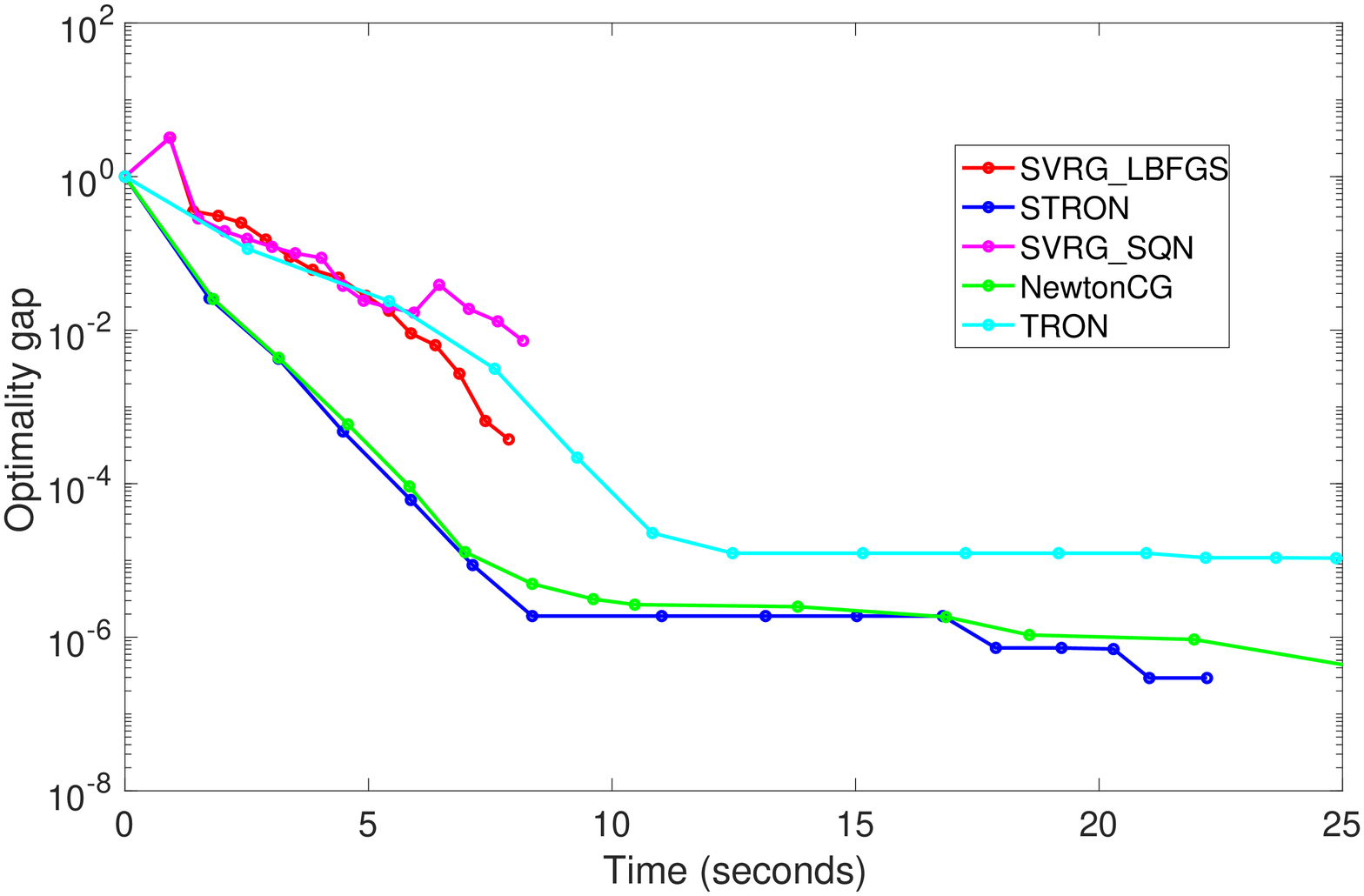}}
	\subfloat{\includegraphics[width=.5\linewidth]{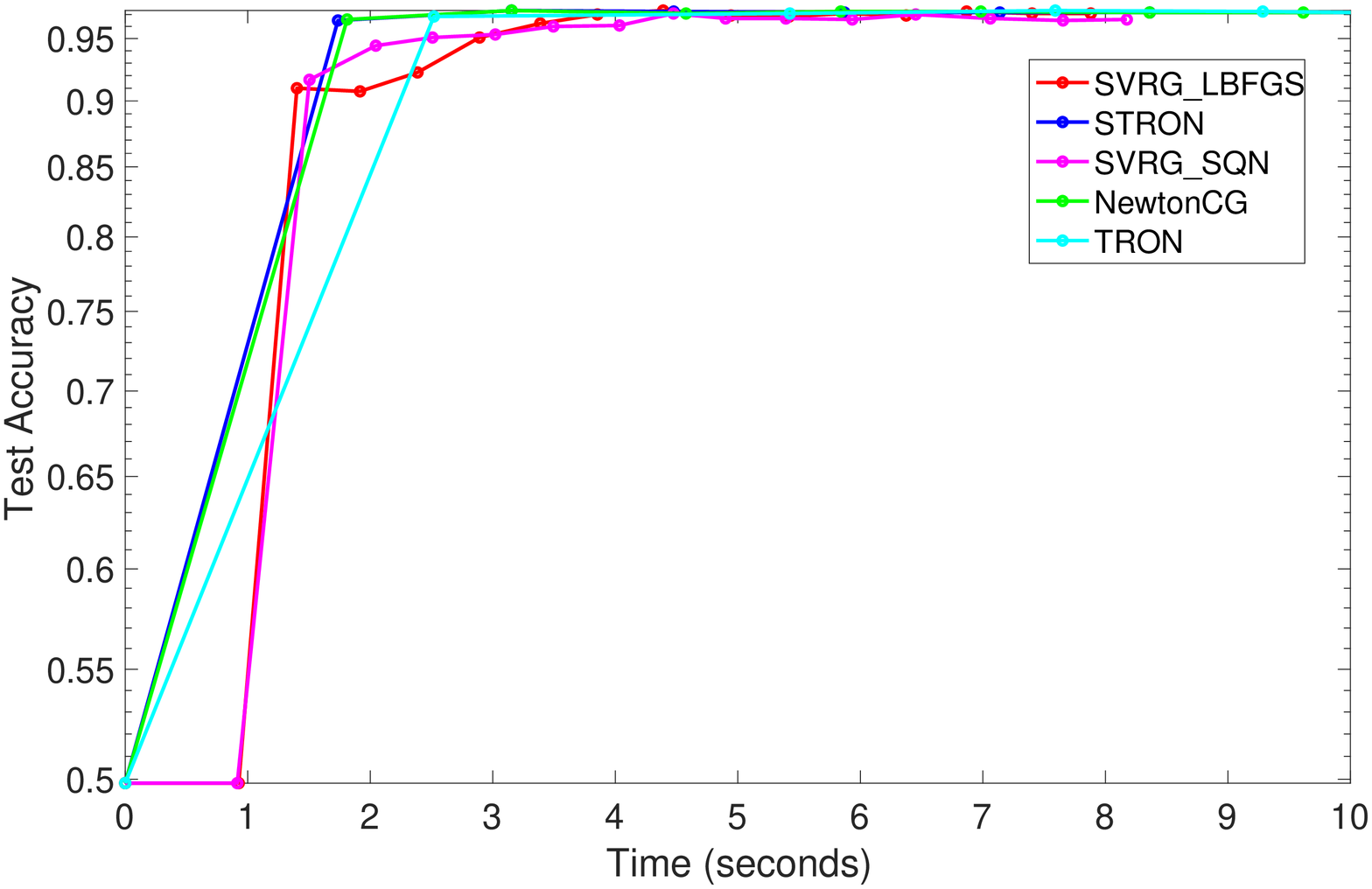}}
	
	\subfloat{\includegraphics[width=.5\linewidth]{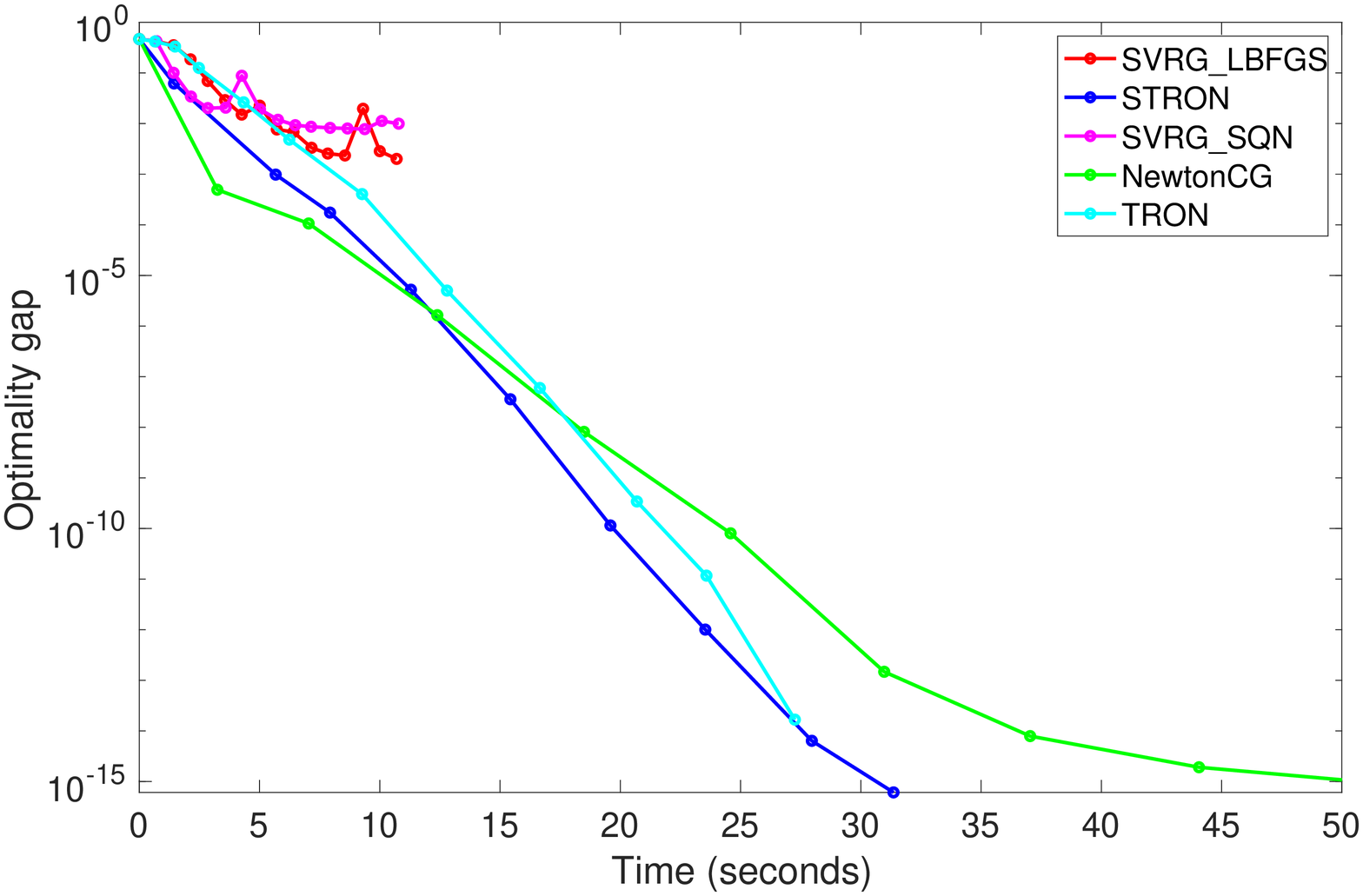}}
	\subfloat{\includegraphics[width=.5\linewidth]{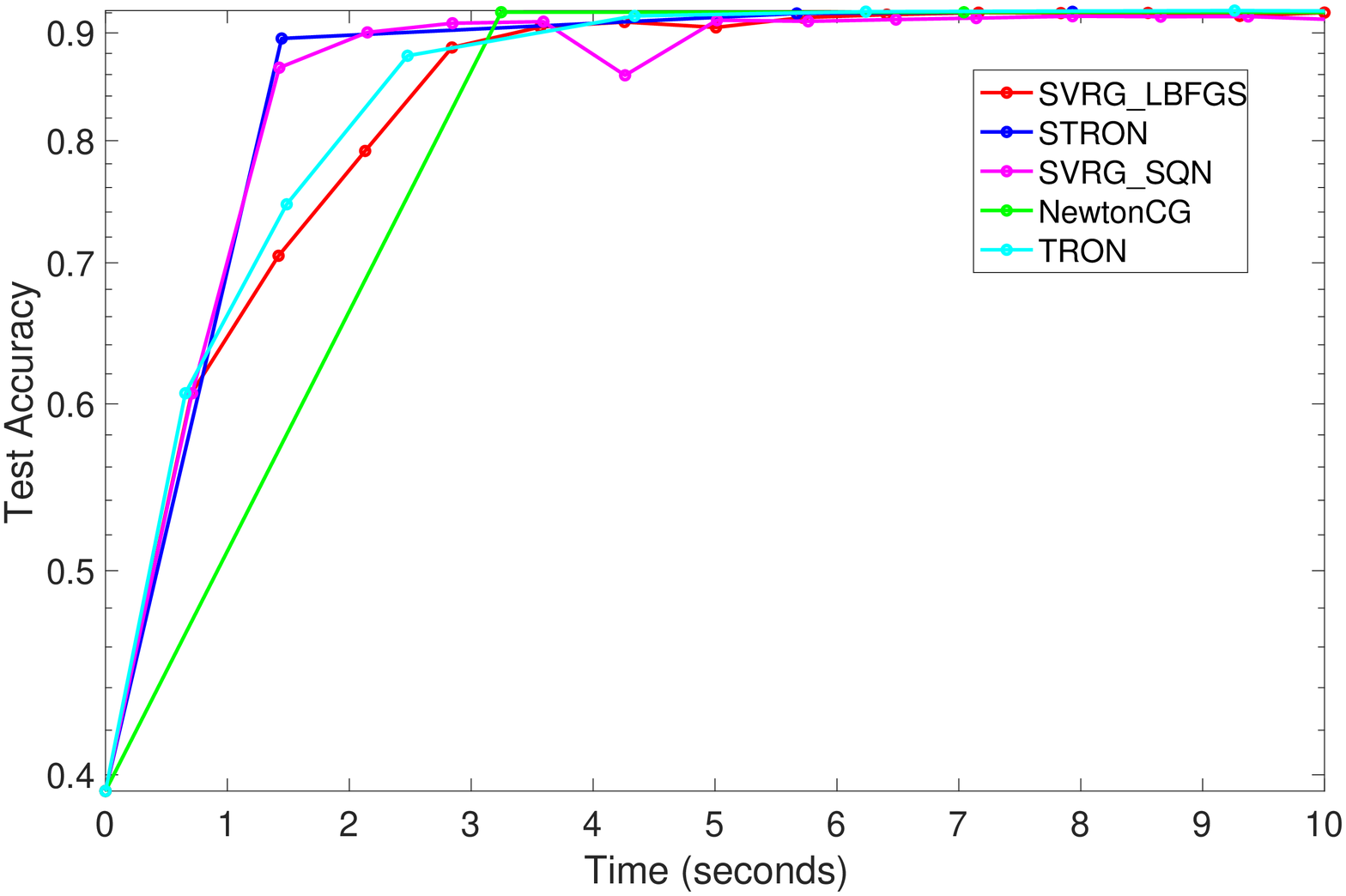}}
	
	\caption{First row presents results with gisette on SVM and second row presents results with webspam on logistic regression.}
	\label{fig_more}
\end{figure}

Fig.~\ref{fig_more} presents more results on gisette and webspam datasets on l$_2$-SVM and l$_2$-regularized logistic regression, respectively. We observe results similar to Figs.~\ref{fig_1}--\ref{fig_svm}, which show that STRON outperforms all other techniques.

\subsection{Effect of Regularization Coefficient}
\begin{figure}[htb]
	\subfloat{\includegraphics[width=.5\linewidth]{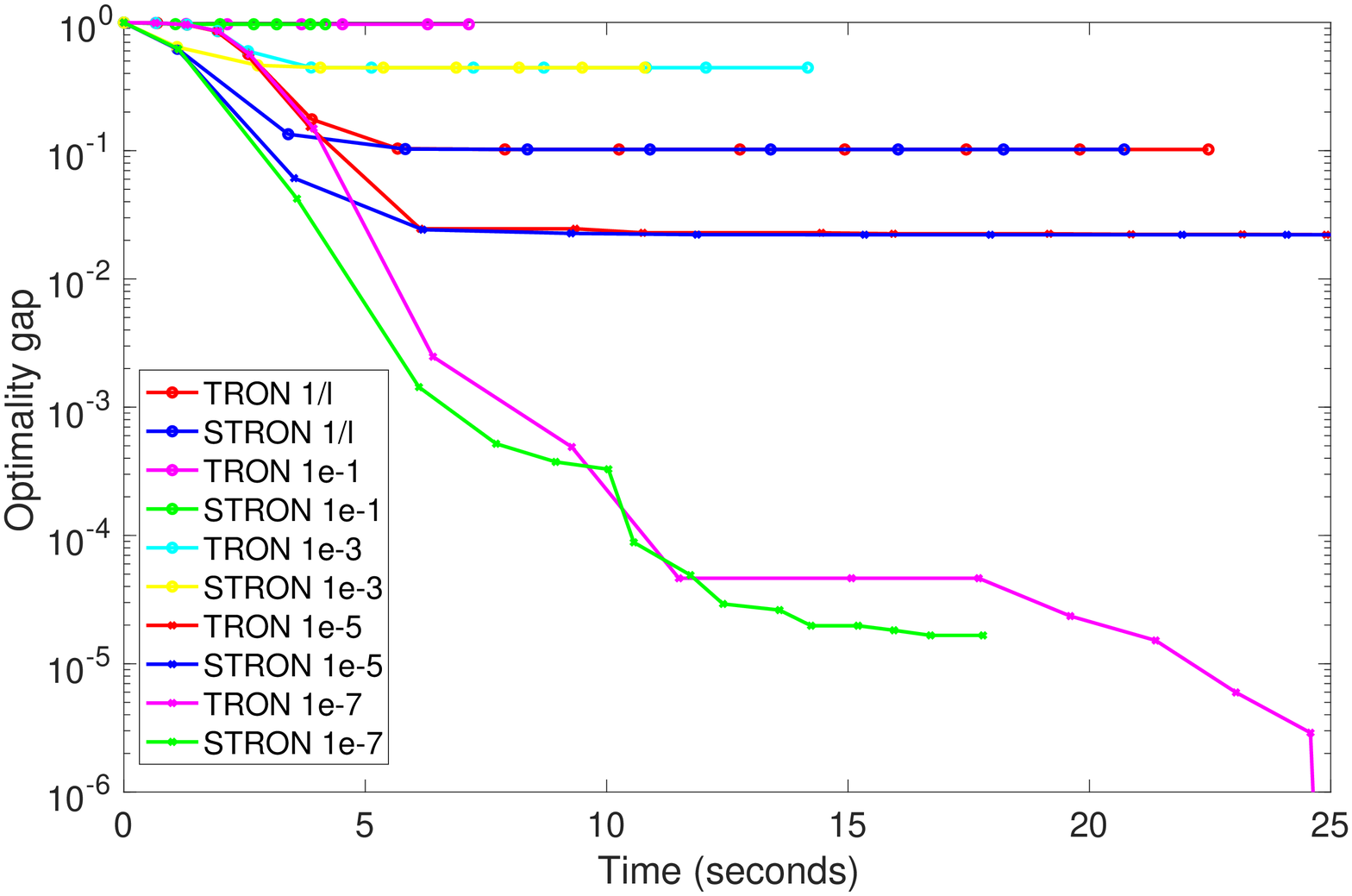}}
	\subfloat{\includegraphics[width=.5\linewidth]{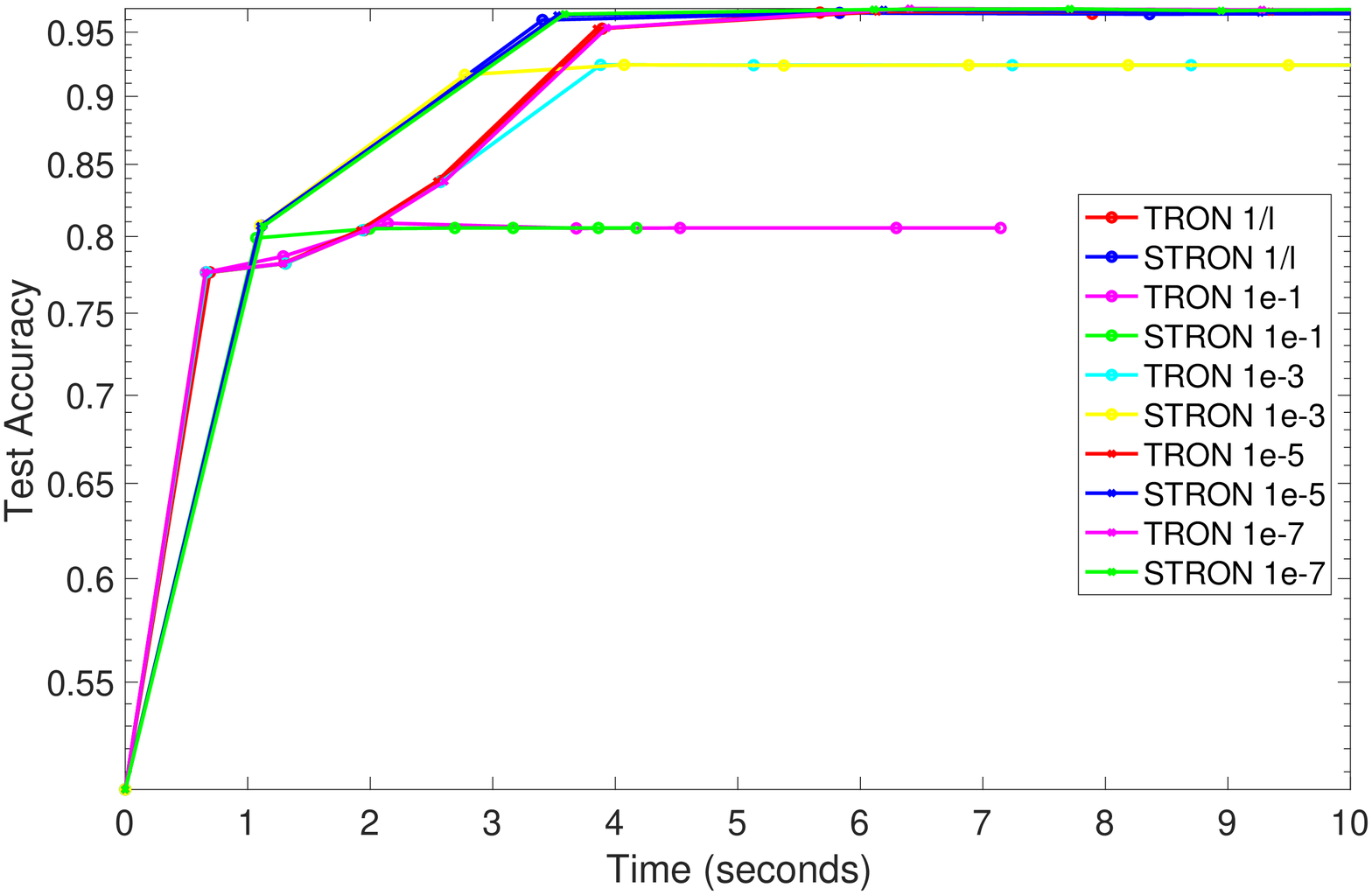}}
	\caption{Effect of regularization coefficient on STRON and TRON methods.}
	\label{fig_lambda}
\end{figure}

Here we study the effect of the value of regularization coefficient ($\lambda$) on the convergence and accuracy of STRON and TRON methods. Fig.~\ref{fig_lambda} presents the results with a series of values for $\lambda~=~\lbrace 1/l, 1e-1, 1e-3, 1e-5, 1e-7 \rbrace$ using news20 dataset with SVM. From the figure, it is clear that both the methods are affected by the choice of value $\lambda$. For larger values of $\lambda~=~\lbrace 1e-1, 1e-3 \rbrace$, both the methods converge to the less accurate solution, as depicted in terms of optimality gap and accuracy plots. On the other hand, for values of $\lambda < 1e-3$, there do not seem be much difference in the accuracy of both the methods. But, in terms of optimality gap, it is clear that smaller the value of $\lambda$, better is the solution, although there seem to be no difference on the corresponding accuracy plot. Moreover, for all the values of $\lambda$, STRON clearly outperforms TRON method.\\
\indent Generally, machine learning problems involve tuning a lot of hyper-parameters which are quite difficult to tune. So to reduce number of parameters to tune, we set $\lambda~= 1/l$, which works well in practice, as it sets the value relative to the size of the dataset and gives good results, as is clear from the figure.

%% BibTeX users please use one of
%\bibliographystyle{spbasic}      % basic style, author-year citations
%%\bibliographystyle{spmpsci}      % mathematics and physical sciences
%%\bibliographystyle{spphys}       % APS-like style for physics
%\bibliography{../../BibDB}   % name your BibTeX data base

% Non-BibTeX users please use

\end{document}